\documentclass[11pt]{article}

\usepackage[english]{babel}

\usepackage[letterpaper,top=1in,bottom=1in,left=1in,right=1in,marginparwidth=1.75cm]{geometry}

\usepackage[utf8]{inputenc} 
\usepackage[T1]{fontenc}    
\usepackage[hidelinks, colorlinks=true,citecolor=blue]{hyperref} 
\usepackage{url}            
\usepackage{booktabs}       
\usepackage[flushleft]{threeparttable}
\usepackage{amsfonts}       
\usepackage{nicefrac}       
\usepackage{microtype}      
\usepackage{amsthm,amsmath,amsfonts, amssymb}
\usepackage{graphicx}
\usepackage{xcolor}
\usepackage{mathrsfs,mathtools}
\usepackage{bm}
\usepackage{authblk}
\usepackage[numbers]{natbib}

\usepackage{caption}
\usepackage{subcaption}
\usepackage{wrapfig}

\usepackage{algorithm}
\usepackage{algorithmic}
\newlength\myindent
\newcommand*{\eg}{\emph{e.g.}{}}
\newcommand*{\ie}{\emph{i.e.}{}}

\title{\bf Conditional Generative Modeling for\\High-dimensional Marked Temporal Point Processes}

\author[1]{Zheng Dong}
\author[2]{Zekai Fan}
\author[2]{Shixiang Zhu\footnote{Email: shixianz@andrew.cmu.edu}}
\affil[1]{{\small Amazon}\vspace{2pt}}
\affil[2]{{\small Carnegie Mellon University}\vspace{-0.3in}}
\date{}

\begin{document}
\maketitle

\begin{abstract}
Point processes offer a versatile framework for sequential event modeling. 
However, the computational challenges and constrained representational power of the existing point process models have impeded their potential for wider applications.
This limitation becomes especially pronounced when dealing with event data that is associated with multi-dimensional or high-dimensional marks such as texts or images.
To address this challenge, this study proposes a novel event-generation framework for modeling point processes with high-dimensional marks. 
We aim to capture the distribution of events without explicitly specifying the conditional intensity or probability density function. Instead, we use a conditional generator that takes the history of events as input and generates the high-quality subsequent event that is likely to occur given the prior observations.
The proposed framework offers a host of benefits, including considerable representational power to capture intricate dynamics in multi- or even high-dimensional event space, as well as exceptional efficiency in learning the model and generating samples. 
Our numerical results demonstrate superior performance compared to other state-of-the-art baselines.
\end{abstract}

\section{Introduction}

Point processes are widely used to model asynchronous event data ubiquitously seen in real-world scenarios, such as earthquakes \citep{ogata1998space, zhu2021imitation}, healthcare records \citep{Dong2023non, Schoenberg2023}, and criminal activities \citep{dong2024atlanta, dong2024spatio, mohler2011self}. These data typically consist of a sequence of timestamps that denote when the events occurred, along with additional descriptive information of events such as category, locations, and even text or image, commonly referred to as ``marks''. With the rise of complex systems, advanced models that go beyond the classic parametric point processes \citep{hawkes1971spectra} are craved to capture intricate dynamics involved in the data-generating mechanism.
Neural point processes \citep{shchur2021neural}, such as Recurrent Marked Temporal Point Processes \citep{du2016recurrent} and Neural Hawkes \citep{mei2017neural}, are powerful methods for event modeling and prediction. They use neural networks (NNs) to model the event intensity and capture complex dependencies among observed events. 

Nonetheless, existing neural point processes face significant challenges when applied to modeling high-dimensional event marks. 
One major challenge falls into model learning.
Current intensity-based methods often learn the model parameters via the commonly used maximum likelihood approach \citep{reinhart2018review}. The computation of point process likelihood involves integrating the event intensity function over time and mark space, which, due to the use of NNs, is often analytically intractable. Therefore, numerical approximations such as Monte Carlo estimation are adopted \citep{dong2023spatiotemporal, mei2017neural, zuo2020transformer} for likelihood estimation. However, complex and extremely expensive numerical approximations are required for estimating integrals over a high-dimensional mark space; otherwise, they introduce large approximation bias and compromise the model accuracy.
Alternative approaches are explored to avoid the numerical integration of the intensity function, such as directly modeling the cumulative intensity function \citep{omi2019fully}, which cannot be defined in high-dimensional space; or adopting tractable parametric forms for the event distribution \citep{shchur2021neural, zhou2022neural}, which restricts the model flexibility and fails to capture complex dynamics among events.

More seriously, these models face pronounced limitations in \emph{generating events with high-dimensional marked information}, as they rely on the event intensity function for event simulation through the thinning algorithm \citep{ogata1981lewis}. The thinning algorithm initially generates large amounts of samples uniformly in the original data space and then rejects the majority of these generated samples according to the predicted conditional intensity. This can be a costly or even impossible task when the mark space is high-dimensional.
As a result, the applicability of these models is significantly limited, particularly in modern applications \citep{williams2020point, zhu2022spatiotemporal}, where event data often come with high-dimensional marks, such as texts and images in police crime reports or social media posts. 

\begin{figure}[!t]
    \centering
    \includegraphics[width=.7\linewidth]{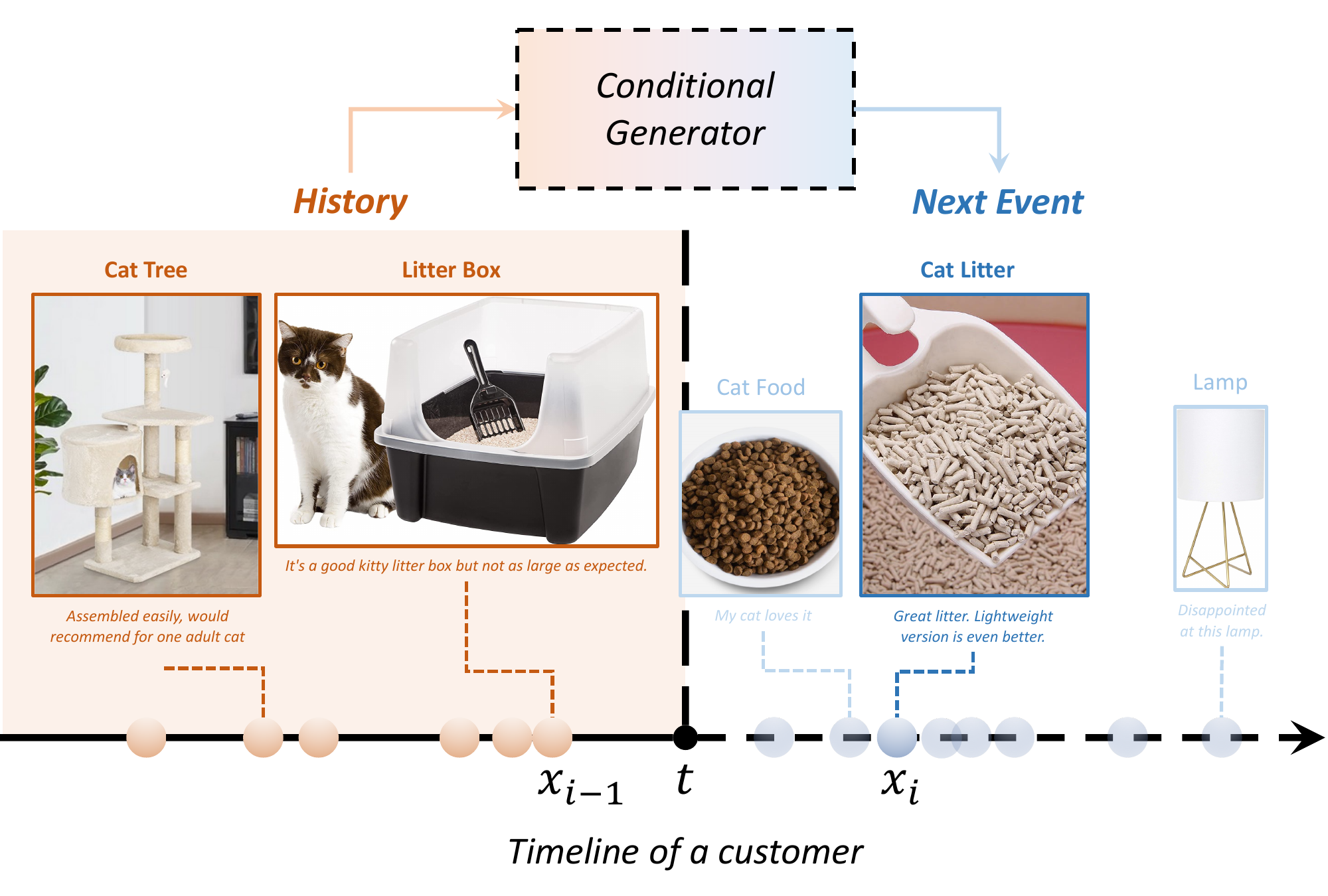}
    \caption{An example of generating high-dimensional content over time. The conditional generator explores the customer's next possible activity, including not only the purchase time, but also the item, and even its image or review. The observed events from the customer's past purchases are represented by yellow dots, while the next generated event is indicated by a blue dot. 
    }
\label{fig:motivating-exp}
\vspace{-0.05in}
\end{figure}

To address the above challenges, this paper presents a novel generative framework for modeling and generating point processes with high-dimensional marks, leveraging recent developments in generative modeling \citep{ho2020denoising, kingma2013auto, sohn2015learning}.
Our approach differs from traditional point process models that rely on estimating the conditional intensity or probability density function to model event distributions \citep{dong2023spatiotemporal, du2016recurrent, mei2017neural, shchur2021neural}. Instead, it uses a conditional event generator to estimate the distribution of future events based on sampling. 
Specifically, the event generator takes the observed history summarized by a history encoder as its input and is able to generate future events following a distribution defined by the model parameters.
The distribution is estimated by approximating the model likelihood using generated and true events, avoiding numerical approximations during model learning.
Moreover, instead of relying on the sampling-rejection procedure in the thinning algorithm, we can jointly simulate future event times and marks that conform to the learned distribution. An illustrative example of event generation is shown in Figure~\ref{fig:motivating-exp}.
In general, the benefits of our model are:
\begin{enumerate}
\item Our model is capable of handling high-dimensional marks such as images or texts, an area not extensively delved into in prior marked point process research;
\item Our model overcomes the computational challenges during both the training phase and the event generation process. 
\item Our model possesses superior representative power, as it does not confine the conditional intensity or probability density of the point process to any specific parametric form;
\item Our model outperforms state-of-the-art baselines in terms of estimation accuracy and generating high-quality event series;
\end{enumerate}
It is important to note that our proposed framework is general and model-agnostic, meaning that a wide spectrum of generative models and learning algorithms can be applied within our framework according to practical needs.
In this paper, we present the instantiation of our framework with a conditional denoising diffusion model as the event generator. We then compare this implementation with other variants, as detailed in the Appendix, through extensive numerical experiments.

\subsection{Related work}

Seminal works \citep{hawkes1971spectra, ogata1998space} introduce self-exciting point processes with parametric kernels. While these models have proven useful, they face limitations in capturing the intricate patterns observed in real-world applications.
Neural point processes like recurrent marked temporal point processes (RMTPP) \citep{du2016recurrent} and Neural Hawkes \citep{mei2017neural} have leveraged recurrent neural networks (RNNs) to embed event history into a hidden state, which then represents the conditional intensity function. 
Another line of research opts for a more ``parametric'' way, which only replaces the parametric kernel in the conditional intensity using neural networks \citep{Dong2023non, okawa2021dynamic, zhu2021imitation, zhu2023sequential}. 
However, these methods can be computationally intractable when dealing with multi-dimensional marks due to the need for numerical integration.
Alternatives have been proposed \citep{chen2021neural, dascher2023using, omi2019fully, Shchur2020Intensity-Free, zhou2022neural} to overcome the challenges, focusing on modeling the cumulative hazard function or conditional probability rather than the conditional intensity, thereby eliminating the need for numerical integration. 
Nonetheless, these methods are used for low-dimensional event data and still rely on the thinning algorithm for event generation that has limited applicability in modern applications. 
Some studies \citep{dong2023spatiotemporal, zhu2022spatiotemporal} only model a simplified and finite high-dimensional mark space, whereas we consider a continuous high-dimensional space. Note that the dimensionality of the mark space is different from its cardinality that previous studies \citep{wu2019learning} refer to, and research on handling mark spaces with high dimensionality is still limited.

Our paper is closely related to the field of generative modeling, which aims to generate high-quality samples from learned data distributions. Prominent models include generative adversarial networks (GANs) \citep{goodfellow2014generative}, variational autoencoders (VAEs) \citep{kingma2013auto}, and diffusion models \citep{ho2020denoising, song2020score}. 
Recent studies have introduced conditional generative models \citep{ho2022classifier, mirza2014conditional, sohn2015learning} that can generate diverse and well-structured outputs based on specific input information, which have found broad applications in areas such as reinforcement learning \citep{ajay2022conditional, li2020anomaly}.
In our work, we adopt a similar technique to consider the event history as contextual information to generate high-quality future events.

The application of generative models to point processes has received limited attention. Three influential papers \citep{li2018learning, sharma2019generative, xiao2018learning} have made significant contributions in this area by using RNN-like models to generate future events. They assume the model's output follows a Gaussian distribution, enabling the exploration of the event space, albeit limiting the representational power of the models. To learn the model, they choose to minimize the ``similarity'' (\eg, Maximum Mean Discrepancy or Wasserstein distance) between the generated and the observed event sequences. It is important to note that these metrics are designed to measure the discrepancy between two distributions in which each data point is assumed to be independent of the others. This approach may not always be applicable to temporal point processes, particularly when the occurrence of future events depends on the historical context. 
A similar concept of modeling point processes using conditional generative models is also explored in other concurrent works \citep{lin2022exploring, yuan2023spatio}. However, their approaches differ from ours in terms of the specific architecture used. They propose a diffusion model with an attention-based encoder, while our framework remains model-agnostic, allowing for greater flexibility in selecting different models. Additionally, their works primarily focus on one-dimensional or spatio-temporal events, and do not account for multi- or high-dimensional marks.

\section{Methodology}
\subsection{Background: Marked temporal point processes}

Marked temporal point processes (MTPPs) \citep{reinhart2018review} consist of a sequence of \emph{discrete events} over time. Each event is associated with a (possibly multi-dimensional) {\it mark} that contains detailed information of the event, such as location, nodal information (if the observations are over networks, such as sensor or social networks), and contextual information (such as token, image, and text descriptions). 
Let $T > 0$ be a fixed time-horizon, and $\mathcal{M} \subseteq \mathbb{R}^d$ be the space of marks. We denote the space of observation as $\mathcal{X} = [0, T) \times \mathcal{M}$ and a data point in the discrete event sequence as
\[
    x = (t, m), \quad t \in [0, T) , \quad m \in \mathcal M,
\]
where $t$ is the event time and $m$ represents the mark. 
Let $N_t$ be the number of events up to time $t < T$ (which is random), and $\mathcal{H}_t := \{x_1 , x_2 , \dots, x_{N_t}\}$ denote historical events.
Let $\mathbb{N}$ be the counting measure on $\mathcal{X}$, \ie, for any measurable $S \subseteq \mathcal{X}$, $\mathbb{N}(S) = |\mathcal{H}_t \cap S|.$ 
For any function $\phi:\mathcal{X}\to\mathbb{R}$, the integral with respect to the counting measure is defined as 
$
    \int_{S} \phi(x) d\mathbb N(x) = \sum_{x_i\in\mathcal H_T\cap S} \phi(x_i).
$

The events' distribution in MTPPs can be characterized via the conditional intensity function $\lambda$, which is defined to be the occurrence rate of events in the marked temporal space $\mathcal{X}$ given the events' history $\mathcal{H}_{t(x)}$, \ie,
\begin{equation}
    \lambda(x| \mathcal{H}_{t(x)} ) = \mathbb{E}\left( d\mathbb{N}(x) | \mathcal{H}_{t(x)} \right) / dx,
    \label{eq:cond-intensity}
\end{equation}
where $t(x)$ extracts the occurrence time of the possible event $x$. Given the conditional intensity function $\lambda$, the corresponding conditional probability density function (PDF) can be written as
\begin{equation}
    f(x| \mathcal{H}_{t(x)} ) = \lambda(x| \mathcal{H}_{t(x)} ) \cdot \exp \left( -\int_{[t_n, t(x)) \times \mathcal{M}} \lambda(u| \mathcal{H}_{t(u)} ) du \right).
    \label{eq:cond-prob}
\end{equation}
where $t_n$ denotes the time of the most recent event that occurred before time $t(x)$ (see the survey \citep{reinhart2018review} for more derivation details). 

Maximum likelihood estimation (MLE) has been the commonly used approach for learning the point process models. The log-likelihood of observing a sequence with $N_T$ events can be expressed in terms of $\lambda$ by the product rule of conditional probability as
\begin{equation}
    \ell(x_1, \dots, x_{N_T}) = \int_{\mathcal X}  \log \lambda(x| \mathcal{H}_{t(x)} ) d \mathbb N(x) -  \int_{\mathcal X}  \lambda(x| \mathcal{H}_{t(x)} ) dx.
    \label{eq:log-likelihood}
\end{equation}
Previous methods aim to find the appropriate modeling of $\lambda$ or $f$ for downstream tasks such as event prediction and generation, while our approach goes beyond the explicit parametrization of $\lambda$ or $f$, as illustrated in the next part. 


\vspace{-0.08in}
\subsection{Conditional event generator}
\label{sec:cond-event-generator}

\begin{figure}[!t]
    \centering
        \begin{subfigure}[h]{.52\linewidth}
        \includegraphics[width=\linewidth]{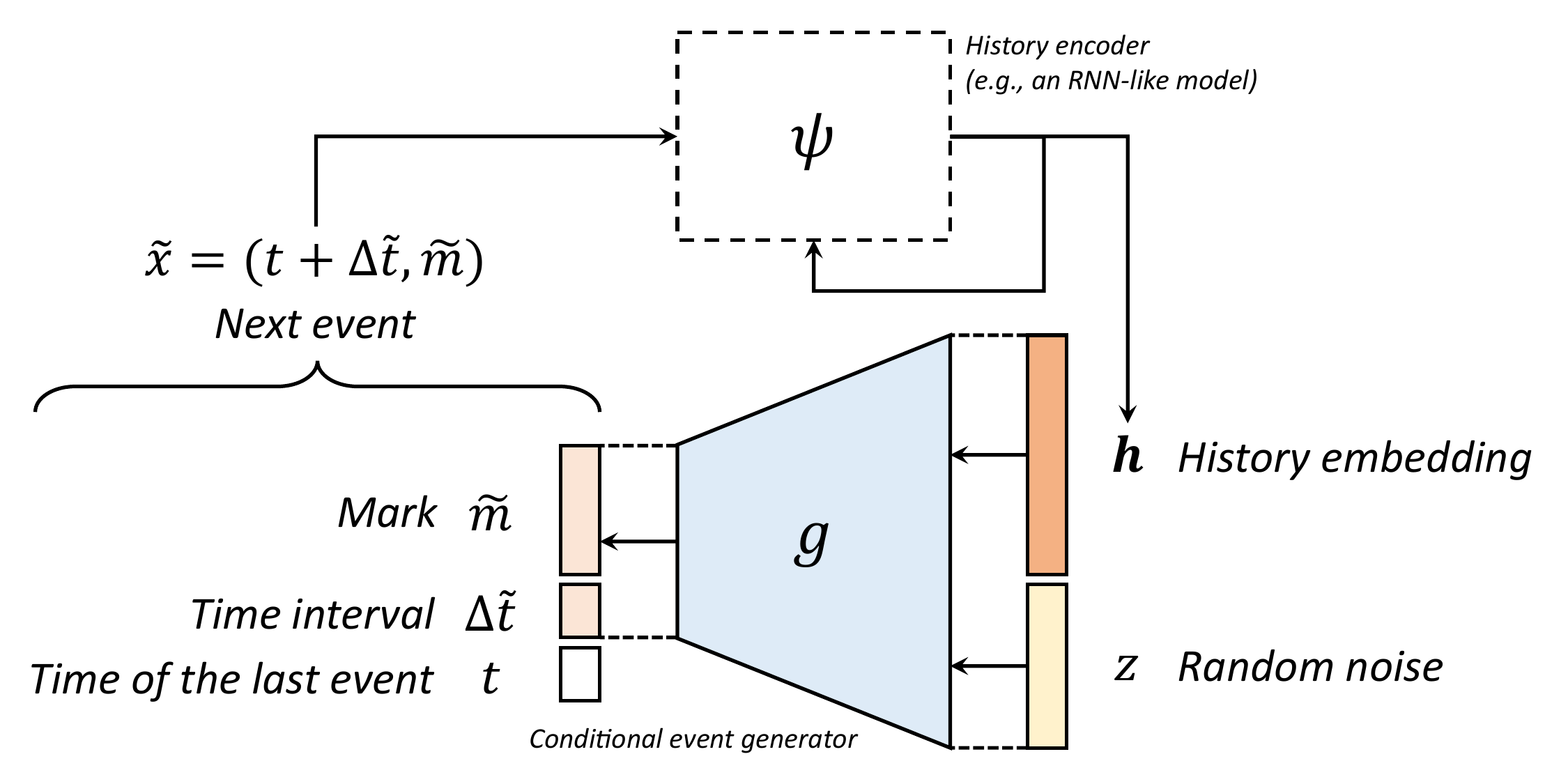}
        \caption{Model architecture}
        \end{subfigure}
    \hspace{0.15in}
    \begin{subfigure}[h]{.40\linewidth}
        \includegraphics[width=\linewidth]{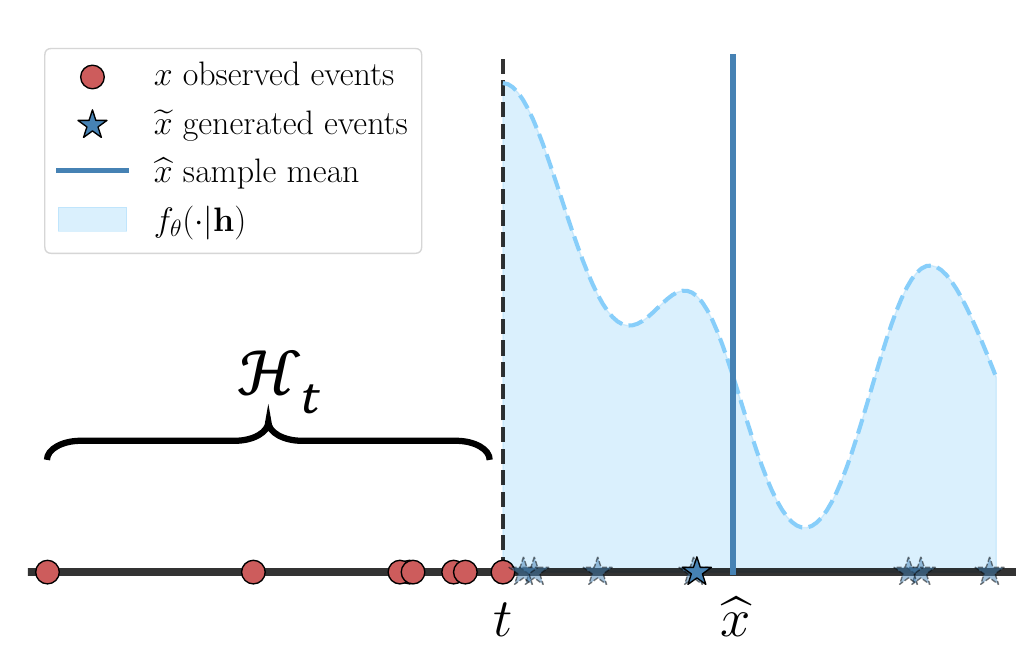}
        \caption{Example of generated events}
    \end{subfigure}
\caption{(a) The architecture of the proposed framework, which consists of two key components: A conditional generative model $g$ that generates $(\Delta \widetilde{t}, \widetilde{m})$ given its history embedding and an RNN-like model $\psi$ that summarizes the events in the history. (b) An example of generated one-dimensional (time only) events $\{\widetilde{x}^{(j)}\}$ given the history $\mathcal{H}_t$. The shaded area suggests the underlying conditional probability density captured by the model parameters $\theta$.
 }
\label{fig:illustration}
\vspace{-.05in}
\end{figure}

The main idea of the proposed framework is to use a \emph{conditional event generator} (\texttt{CEG}) to produce the $i$-th event $x_i = (t_{i-1} + \Delta t_i, m_i)$ given its previous $i-1$ events, without relying on $\lambda$ or $f$. Here, $\Delta t_i$ and $m_i$ indicate the time interval between the $i$-th event and its preceding event and the mark of the $i$-th event, respectively. 
Formally, this is achieved by a generator function:
\begin{equation}
    g(z, \boldsymbol{h}_{i-1}): \mathbb{R}^{r+p} \to (0, +\infty) \times \mathcal{M},
\end{equation}
which takes an input in the form of a random noise vector ($z \in \mathbb{R}^r \sim \mathcal{N}(0, I)$) and a history embedding ($\boldsymbol{h}_{i-1}\in \mathbb{R}^p$) that summarizes the history information up to and excluding the $i$-th event, namely, $\mathcal{H}_{t_i} = \{x_1, \dots, x_{i-1}\}$. 
The output of the generator includes the event time interval and the mark of the $i$-th event denoted by $\Delta \widetilde{t}_i$ and $\widetilde{m}_i$, respectively.
To ensure that the time interval is positive, we restrict $\Delta \widetilde{t}_i$ to be greater than zero.  
It is worth noting that our selection of the generator $g$ is not constrained to a particular function. Instead, it can represent an arbitrary generative process, such as a diffusion model, provided it has the capability to produce a sample when supplied with noise and a history embedding.

To represent the conditioning variable $\boldsymbol{h}_{i-1}$, we use a \emph{history encoder} represented by $\psi$, which has a recursive structure such as recurrent neural networks (RNNs) or Transformers. In our numerical results, we opt for long short-term memory (LSTM), which takes the current event $x_i$ and the preceding history embedding $\boldsymbol{h}_{i-1}$ as input and generates the new history embedding $\boldsymbol{h}_i$. 
This new history embedding represents an updated summary of the past events including $x_i$. Formally, 
\[
    \boldsymbol{h}_0 = \boldsymbol{0}~\text{and}~\boldsymbol{h}_i = \psi(x_i, \boldsymbol{h}_{i-1}),\quad i=1,2,\dots,N_T.
\]
We denote the parameters of both models $g$ and $\psi$ using $\theta \in \Theta$.
The model architecture is presented in Figure~\ref{fig:illustration} (a).

\vspace{-.1in}
\paragraph{Connection to marked temporal point processes}
The proposed framework draws its statistical inspiration from MTPPs. 
Unlike other recent attempts at modeling point processes, our framework \emph{approximates the conditional probability of events using generated samples} rather than directly modeling the conditional intensity in (\ref{eq:cond-intensity}) or PDF in (\ref{eq:cond-prob}) \citep{du2016recurrent, mei2017neural, omi2019fully, Shchur2020Intensity-Free, zhu2022neural, zhu2021deep}. 

Let $\boldsymbol{h}$ denote the history embedding of $\mathcal{H}_{t}$. 
Given $\boldsymbol{h}$, our model generates the subsequent event occuring after $t$, denoted by $\widetilde{x} = (t+\Delta \widetilde{t}, \widetilde{m})$. 
The generated event adheres to the following conditional probabilistic distribution:
\[
    \widetilde{x} \sim f_\theta(\cdot|\boldsymbol{h}) \approx f(\cdot|\mathcal{H}_{t}),
\]
where $f_\theta$ denotes the conditional PDF of the underlying MTPP model (\ref{eq:cond-prob}).
This design has three main advantages compared to other point process models:
\begin{enumerate}
    \item \emph{Generative efficiency}: 
    Our model's generative capability enables highly efficient simulation of event series for any point processes without the need for thinning algorithms \citep{ogata1981lewis}. Thinning algorithms typically generate a large number of random samples ($\mathcal{O}(N^d)$) across a high-dimensional space, only to retain a small portion of them ($\mathcal{O}(N_T)$) based on predicted conditional intensity, where $N_T \ll N^d$. 
    Our method, however, bypasses this by directly generating samples through \texttt{CEG}, significantly reducing complexity from $\mathcal{O}(N^d)$ to $\mathcal{O}(N_T)$. It eliminates the inefficiency incurred by the sample rejection, as shown in Algorithm~\ref{alg:CEG-data-generation}.
    \item \emph{Expressiveness}: The proposed model enjoys considerable representational power, as it does not impose any restrictions on the form of the conditional intensity $\lambda$ or PDF $f$. The numerical findings also indicate that our model is capable of capturing complex event interactions, even in a multi-dimensional space.
    \item \emph{Predictive efficiency}:
    To predict the next event $\widehat{x}_i = (t_{i-1} + \Delta \widehat{t}_i, \widehat{m}_i)$ given the observed events' history $\mathcal{H}_{t_{i}}$, we can calculate the sample average over a set of generated events $\{\widetilde{x}_i^{(l)}\}$ without the need for an explicit expectation computation, \ie,
    \[
        \widehat{x}_i = \int_{(t_{i-1}, +\infty) \times \mathcal{M}} x \cdot f(x | \mathcal{H}_{t(x)}) dx \approx \frac{1}{L} \sum_{l=1}^L~\widetilde{x}_i^{(l)},
    \]
    where $L$ denotes the number of samples. 
\end{enumerate}



\begin{algorithm}[!t]
\begin{algorithmic}
    \STATE {\bfseries Input:} Generator $g$, history encoder $\psi$, time horizon $T$\;
    \STATE {\bfseries Initialization:} $\mathcal{H}_T = \emptyset, \boldsymbol{h}_0 = \boldsymbol{0}, t=0, i=0$\; 
    \WHILE{$t<T$}
        \STATE 1. Sample $z \sim \mathcal{N}(0, I)$;
        \STATE 2. Generate next event $\widetilde{x} = (t + \Delta\widetilde{t}, \widetilde{m})$, where $(\Delta\widetilde{t}, \widetilde{m}) = g(z, \boldsymbol{h}_i)$;
        \STATE 3. $i = i+1$; $t = t + \Delta\widetilde{t}; x_i = \widetilde{x}$; $\mathcal{H}_T = \mathcal{H}_T \cup \{x_i\}$; 
        \STATE 4. Update history embedding $\boldsymbol{h}_i = \psi(x_i, \boldsymbol{h}_{i-1})$;
    \ENDWHILE
    \IF{$t(x_i) \geq T$}
        \STATE {\bfseries return} $\mathcal{H}_T - \{x_i\}$  \;
    \ELSE
        \STATE {\bfseries return} $\mathcal{H}_T$ \;
    \ENDIF
\end{algorithmic}
\caption{Event generation process using \texttt{CEG}}
\label{alg:CEG-data-generation}
\end{algorithm}

\subsection{Model estimation}
\label{sec:model-estimation}

To learn the model, one can maximize the log-likelihood of the observed event series. 
Instead of directly specifying the log-likelihood using $\lambda$ as in \eqref{eq:log-likelihood}, an equivalent form of this objective can be expressed using conditional PDF, as shown in the following equation (see Appendix~\ref{append:derivation-cond-prob}):
\begin{equation}
\max_{\theta \in \Theta}~ \ell(\theta) \coloneqq \frac{1}{E} \sum_{e=1}^E \int_\mathcal{X} ~\log f_\theta(x|\boldsymbol{h})~d\mathbb{N}_e(x),
\label{eq:sequence-likelihood}
\end{equation}
where $E$ represents the total number of observed event sequences and $\mathbb{N}_e$ is the counting measure of the $e$-th event sequence. Therefore, The log-likelihood \eqref{eq:log-likelihood} is the summation of the log conditional probabilistic density function $f_{\theta}$ evaluated at all events. 
It is worth noting that this learning objective circumvents the need to compute the intensity function and its integral in the second term of \eqref{eq:log-likelihood}, a task that can be computationally intractable when events exist in a multi- or high-dimensional data space.


Now the key challenge is \emph{how do we maximize the model likelihood without access to the function $f_\theta$}? 
This is a commonly posed inquiry in the realm of generative model learning, and there are several pre-existing learning algorithms intended for generative models that can provide solutions to this question \citep{bond2021deep}. 
In the rest of this section, we present the learning strategy of denoising score matching by choosing a conditional diffusion model as the CEG in our framework.
In this way, a denoising process is assumed to transform a noise distribution to the real data distribution, and we can use the generated samples to match the denoising scores of this process through a score function parametrized by a neural network, therefore maximizing the log-likelihood of the real data \citep{ho2020denoising, song2019generative}.
Other learning algorithms for likelihood maximization can also be adopted, and we provide two additional approaches in Appendix~\ref{app:non-param-learning} and \ref{app:variational-learning}, which serves as ablation study in our experiments.

\vspace{-0.1in}
\paragraph{Conditional denoising diffusion model}
For ease of presentation, we only consider the conditional PDF of a single event $x$ in the following derivation.
The conditional denoising diffusion model (\texttt{CDDM}) approximates the data distribution in the form of $f_{\theta}(x|\bm{h}) \coloneqq \int p_{\theta}(x, z_{1:K}|\bm{h})dz_{1:K}$, where $z_{1:K}$ are latent variables of the same dimensionality as the data $x$ and $p_{\theta}$ is the model to be learned.
The variational distribution of the latent variables $q(z_{1:K}|x, \bm{h})$ is assumed by a forward process, which is fixed to a Markov chain to gradually add noise to the data $x$ with variance schedule $\{\beta_k\}_{k=1}^{K}$. Another reverse process inverts the transformation from the noise distribution $\mathcal{N}(0, I)$ back to data, with Gaussian transitions parametrized by the model parameters $\theta$. The model can be learned by maximizing the variational bound on the log-likelihood of the data $x$:
\[
    \begin{aligned}
        \log f_{\theta}(x|\bm{h}) \geq \mathbb{E}_q\left[ \log p(z_K|\bm{h}) + \log\frac{p_{\theta}(x|z_1, \bm{h})}{q(z_1|x, \bm{h})} \right. + \left. \sum_{k>1}\log\frac{p_{\theta}(z_{k-1}|z_k, \bm{h})}{q(z_k|z_{k-1}, \bm{h})} \right].
    \end{aligned}
\]
The variational bound can be expressed by the KL divergence between a series of Gaussian distributions of $z_{1:K}$ and $x$, leading to the following equivalent objective:
\[
    \mathbb{E}_q\left[ \sum_{k>1}\frac{1}{2\beta_k}\|\tilde{\mu}_k(z_k, x) - \mu_{\theta}(z_k, k | \bm{h})\|^2 \right] + C,
\]
where $\mu_{\theta}$ is the mean function of the Gaussian transitions to be learned. The $\tilde{\mu}_k$ is the mean of forward process posteriors for the reverse process to match, with $\tilde{\mu}_k(z_k, x) = \frac{\sqrt{\bar{\alpha}_{k-1}}\beta_k}{1 - \bar{\alpha}_k}x + \frac{\sqrt{\alpha_k}(1 - \bar{\alpha}_{k-1})}{1 - \bar{\alpha}_k}z_k$, $\alpha_k \coloneqq 1-\beta_k, \bar{\alpha}_k \coloneqq \prod_{r=1}^k\alpha_r$, and $C$ is a constant independent with $\theta$. Using $\epsilon$-prediction \citep{ho2020denoising}, we can finally train the model by
\begin{equation}
    \min_{\theta \in \Theta}~ \mathbb{E}_{k, \epsilon}\left[ \| \epsilon - \epsilon_{\theta}(\sqrt{\bar{\alpha}_k}x + \sqrt{1-\bar{\alpha}_k}\epsilon, \bm{h}, k)\|^2 \right].
    \label{eq:score-matching-loss}
\end{equation}
The $\epsilon_{\theta}$ is the score function parametrized by a neural network for denoising score matching, and the learned score $\epsilon_{\theta}(\sqrt{\bar{\alpha}_k}x + \sqrt{1-\bar{\alpha}_k}\epsilon, \bm{h}, k)$ estimates the gradient of the log-density of the conditional distribution of $z_k$ using generated samples. The final loss function is composed of the score-matching loss at all observed events (\ie, summing \eqref{eq:score-matching-loss} over $x$ and corresponding $\bm{h}$). 

\begin{algorithm}[!t]
\begin{algorithmic}
    \STATE {\bfseries Input:} noise $z_K \sim \mathcal{N}(0, I)$, history embedding $\bm{h}$\;
    \FOR{$k=K, \dots, 1$}
        \STATE 1. Sample $\epsilon \sim \mathcal{N}(0, I)$ if $k>1$, else $\epsilon = 0$;
        \STATE 2. $\tilde{\epsilon} = (1+w)\epsilon_{\theta}(z_k, \bm{h}, k) - w\epsilon_{\theta}(z_k,\varnothing,k)$, where $w$ is the pre-chosen guidance strength;
        \STATE 3. $z_{k-1} = \frac{1}{\sqrt{\alpha_t}}\left(z_k - \frac{1-\alpha_t}{\sqrt{1-\bar{\alpha}_t}}\tilde{\epsilon}\right) + \sqrt{\beta}_t \epsilon$;
    \ENDFOR
    \STATE {\bfseries return} $\widetilde{x} \coloneqq z_0$  \;
\end{algorithmic}
\caption{Generator $g$ in \texttt{CDDM} with classifier-free guidance}
\label{alg:generator-g}
\end{algorithm}

We choose the classifier-free diffusion guidance \citep{ho2022classifier} to achieve the conditional sampling using the denoising diffusion model. 
By incorporating this model architecture, the generator function $g$ is chosen to be the entire reverse process for sampling data from any given noise, as illustrated in Algorithm~\ref{alg:generator-g}.
See Appendix~\ref{app:conditional-ddpm} for derivation and implementation details of training and sampling.

\subsection{Alternative variants}

We present two variants of the proposed framework, choosing the conditional event generator as one of the other deep learning architectures. These variants, being evaluated in experiments, demonstrate the broad applicability of our framework, as the event generator can be flexibly chosen to adapt to different scenarios.

\vspace{-0.1in}
\paragraph{Non-parametric learning} In low-dimensional (1D or 2D) cases, the conditional PDF in \eqref{eq:sequence-likelihood} can be estimated through kernel density estimation (KDE) using generated samples.
Given the training data $x$ and the history embedding $\bm{h}$, we first generate a set of candidates $\{\widetilde{x}^{(l)}\}_{l=1}^L$ via $g_{\theta}(\cdot , \boldsymbol{h})$. The conditional PDF at $x$ can be estimated by
\begin{equation}
    f_{\theta}(x| \bm{h} ) \approx \frac{1}{L} \sum_{l=1}^L \kappa_\sigma (x - \widetilde{x}^{(l)}),
    \label{eq:kde}
\end{equation}
where $\kappa_\sigma$ is a kernel function with a bandwidth $\sigma$. The complete model log-likelihood is estimated using KDE at each observed event given the corresponding history. 

Two main challenges exist when estimating $f(x|\bm{h})$ via KDE: ($i$) The density of events can vary from location to location and may also change significantly over the training iterations. Consequently, using a single bandwidth for estimation would either oversmooth the conditional PDF or introduce excessive noise in areas with sparse events.
($ii$) The support of the next event's time is $[0,+\infty)$, and the time interval for the next event to come can cluster in a small neighborhood near $0$, which will lead to a significant boundary bias.
To this end, we adopt the self-tuned kernel \citep{cheng2022convergence, mall2013self} with boundary correction \citep{jones1993simple} by choosing the bandwidth adaptively when evaluating the density at different data points, and correcting the boundary bias by reflecting the data points against $0$ in the time domain. 
Figure~\ref{fig:example-of-kde} shows the benefits of using the self-tuned kernel with boundary correction for KDE. See more details in Appendix~\ref{app:non-param-learning}.

\vspace{-0.1in}
\paragraph{Variational learning} The popular variational method for learning generative models can also be adopted in our framework.
Following the idea of conditional variational autoencoder (CVAE) \citep{sohn2015learning}, we approximate the log conditional PDF using its evidence lower bound (ELBO) (see derivation details in Appendix~\ref{app:variational-learning}):
\begin{equation}
    \begin{aligned}
        \log f_{\theta}(x|\bm{h}) &\geq -D_\text{KL}(q(z|x, \boldsymbol{h})||p_{\theta}(z|\boldsymbol{h})) + \mathbb{E}_{q(z|x, \boldsymbol{h})}\left[\log p_{\theta}(x|z, \boldsymbol{h}) \right],
    \end{aligned}
    \label{eq:variational-lower-bound} 
\end{equation}
where $q$ is a variational approximation of the posterior distribution of noise $z$ given observed event $x$ and its history $\boldsymbol{h}$.
The first term on the right is the Kullback–Leibler (KL) divergence of the approximate posterior $q(\cdot|x, \boldsymbol{h})$ from the exact posterior $p_{\theta}(\cdot|\boldsymbol{h})$.
The second term is the log-likelihood of the latent data-generating process. 

When using variational learning, Both $q(z|x, \boldsymbol{h})$ and $p_{\theta}(z|\boldsymbol{h})$ are modeled as Gaussian distributions, which allows us to compute the KL divergence with a closed-form expression. 
A common choice for $q(z|x, \boldsymbol{h})$ is a factorized Gaussian distribution:
\[
    q(z|x, \boldsymbol{h}) = \mathcal{N}(z; \mu, \text{diag}(\Sigma)) = \prod_{j=1}^r \mathcal{N}(z_j; \mu_j, \sigma_j^2).
\]
The $\mu = (\mu_j)_{j=1,\dots,r}$ and $\text{diag}(\Sigma) = (\sigma_j^2)_{j=1,\dots,r}$ are the output of a \emph{encoder net} $g_\text{encode}(\epsilon, x, \boldsymbol{h})$ (a fully-connected neural network) and the latent variable $z$ can be obtained using reparametrization trick:
$
    z = \mu + \text{diag}(\Sigma) \odot \epsilon, 
$
where $\epsilon \sim \mathcal{N}(0, I)$ is a random variable and $\odot$ is the element-wise product. The $p_{\theta}(z|\boldsymbol{h})$ can be relaxed as a standard Gaussian distribution $p_\theta(z)$ \cite{kingma2014semi, sohn2015learning}.
The maximization of the second term is equivalent to maximizing the cross entropy between the distribution of an observed event $x$ and the reconstructed event $\widetilde{x}$ by a decoder given $z$ and $\boldsymbol{h}$. Here, our event generator acts as the decoder. Thus, the second term is implemented as the reconstruction loss between $x$ and $\widetilde{x}$.

\section{Experiments}

We evaluate our method using both synthetic and real data and demonstrate its superior performance compared to six state-of-the-art approaches, including (1) Recurrent marked temporal point processes (\texttt{RMTPP}) \cite{du2016recurrent}, (2) Neural Hawkes (\texttt{NH}) \citep{mei2017neural}, (3) Transformer Hawkes process (\texttt{THP}) \citep{zuo2020transformer}, (4) Fully neural network based model (\texttt{FullyNN}) \citep{omi2019fully}, (5) Epidemic type aftershock sequence (\texttt{ETAS}) \citep{ogata1998space} model, (6) Deep non-stationary kernel in point process (\texttt{DNSK}) \citep{dong2023spatiotemporal}. The first four baselines leverage neural networks to model temporal event data (or only with categorical marks). The last two baselines are chosen for testing multi- and high-dimensional event data. Meanwhile, the \texttt{DNSK} is the state-of-the-art method that uses neural networks for high-dimensional mark modeling. See a survey of neural temporal point processes \cite{shchur2021neural} for a detailed review of baselines\footnote{Code implementation of our method is available at {\color{magenta}\url{https://github.com/McDaniel7/Conditional_Generative_Point_Processes/tree/main}}.}.

\begin{table*}[!t]
  \caption{Performance comparison with five baseline methods (bold indicates the best performance).}
  \vspace{-.1in}
  \centering
  \resizebox{1.\linewidth}{!}{
  \begin{threeparttable}
  \begin{tabular}{ccccccccccc}
    \toprule
    \toprule
    & \multicolumn{3}{c}{\bf 1D self-exciting data} & \multicolumn{3}{c}{\bf 1D self-correcting data} & \multicolumn{3}{c}{\bf 3D synthetic data } & {\bf 3D earthquake data}\\
    \cmidrule(lr){2-4} \cmidrule(lr){5-7} \cmidrule(lr){8-10} \cmidrule(lr){11-11}
    Model & Testing $\ell$ ($\uparrow$) & MRE of $f$ ($\downarrow$) & MRE of $\lambda$ ($\downarrow$) & Testing $\ell$ & MRE of $f$ & MRE of $\lambda$ & Testing $\ell$ & MRE of $f$ & MRE of $\lambda$ & Testing $\ell$ \\
    \midrule
    \texttt{RMTPP} & $-1.051~(0.015)$ & $0.437$ & $0.447$ & $-0.975~(0.006)$ & $0.308$ & $0.391$ & / & / & / & / \\
    \texttt{NH} & $-0.776~(0.035)$ & $0.175$ & $0.198$ & $-1.004~(0.010)$ & $0.260$ & $0.363$ & / & / & / & / \\
    \texttt{THP} & $-0.718~(0.013)$ & $0.081$ & $0.103$ & $-0.926~(0.018)$ & $0.169$ & $0.202$ & / & / & / & / \\
    \texttt{FullyNN} & $-0.729~(0.003)$ & $0.083$ & $0.105$ & $-0.821~(0.008)$ & $0.121$ & $0.167$ & / & / & / & / \\
    \texttt{ETAS} & / & / & / & / & / & / & $-4.832~(0.002)$ & $0.981$ & $0.902$ & $-3.939~(0.002)$\\
    \texttt{DNSK} & $-0.649~(0.002)$ & $0.015$ & $\textbf{0.024}$ & $-2.832~(0.004)$ & $0.134$ & $0.185$ & $-2.560~(0.004)$ & $0.339$ & $0.415$ & $-3.606~(0.003)$\\
    \midrule
    \texttt{CEG+KDE} & $\textbf{--0.645}~(0.002)$ & $\textbf{0.013}$ & $0.066$ & $\textbf{--0.768}~(0.005)$ & $\textbf{0.042}$ & $\textbf{0.075}$ & $-2.540~(0.011)$ & $0.056$ & $0.101$ & $-2.629~(0.015)$ \\
    \texttt{CEG+CVAE} & $-0.702~(0.006)$ & $0.073$ & $0.102$ & $-0.941~(0.010)$ & $0.165$ & $0.379$ & $-2.558~(0.009)$ & $0.093$ & $0.156$ & $-3.505~(0.023)$\\
    \texttt{CEG+CDDM} & $-0.691~(0.004)$ & $0.067$ & $0.089$ & $-0.775~(0.003)$ & $0.117$ & $0.187$ & $\textbf{--2.501}~(0.008)$ & $\textbf{0.042}$ & $\textbf{0.097}$ & $\textbf{--0.299}~(0.006)$ \\
    \texttt{CEG+TF+CDDM} & $-0.658~(0.003)$ & $0.044$ & $0.073$ & $-0.770~(0.003)$ & $0.098$ & $0.156$ & $\textbf{--2.502}~(0.004)$ & $0.043$ & $0.101$ & $\textbf{--0.297}~(0.005)$ \\
    \bottomrule
    \bottomrule
  \end{tabular}
  \begin{tablenotes}
  \item *Numbers in parentheses present standard error for three independent runs.
  \end{tablenotes}
  \end{threeparttable}
  }
  \label{tab:synthetic-data-results}
\end{table*}

\subsection{Synthetic data}

We first evaluate our model's performance on synthetic data when equipped with a variety of architectures, aiming to demonstrate the effectiveness of the model-agnostic framework in recovering the ground truth (\ie, the conditional distribution or intensity from the observed events).
Three variants of our model adopt an LSTM as the history encoder and choose the conditional event generator as a fully-connected neural network (\texttt{CEG+KDE}, see Appendix~\ref{app:non-param-learning}), a conditional variational auto-encoder (\texttt{CEG+CVAE}, see Appendix~\ref{app:variational-learning}), and a conditional denoising diffusion model (\texttt{CEG+CDDM}, our main architecture). We also showcase the model's effectiveness with the history encoder being a Transformer (\texttt{CEG+TF+CDDM}) using quantitative evaluation. Details about the experimental setup and our model architecture are presented in Appendix~\ref{append:additional-results}.

\begin{figure}[!t]
\centering
    \begin{subfigure}[h]{.48\linewidth}
        {\includegraphics[width=.49\linewidth]{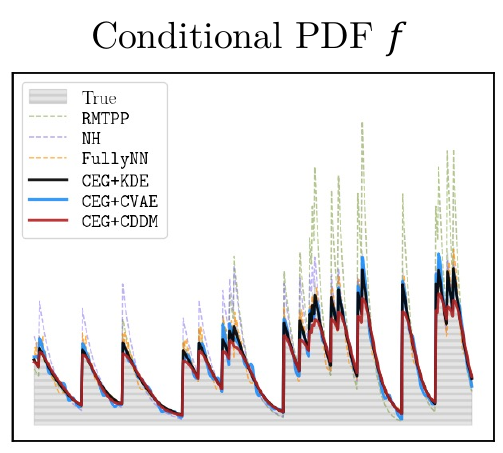}}
        {\includegraphics[width=.49\linewidth]{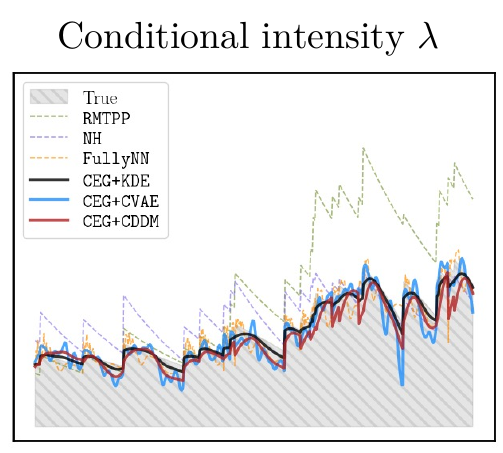}}
        \caption{Self-exciting}
    \end{subfigure}
    \begin{subfigure}[h]{.48\linewidth}
        {\includegraphics[width=.49\linewidth]{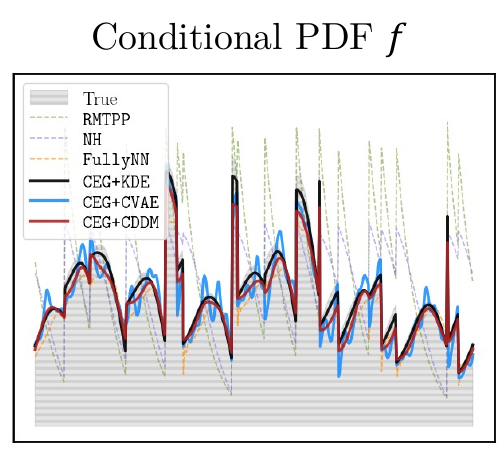}}
        {\includegraphics[width=.49\linewidth]{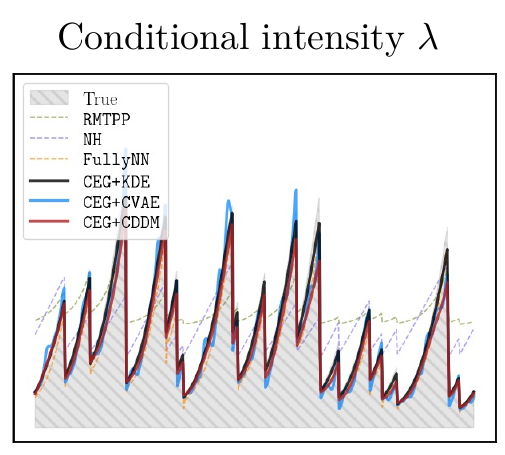}}
        \caption{Self-correcting}
    \end{subfigure}
\caption{Out-of-sample estimation of the conditional PDF $f(t|\mathcal{H}_t)$ and the intensity $\lambda(t|\mathcal{H}_t)$ on one-dimensional (time only) synthetic data sets. One sequence is picked from each testing set for evaluation. The grey shaded areas represent the true $f(t|\mathcal{H}_t)$ and the true $\lambda(t|\mathcal{H}_t)$.}
\vspace{-.05in}
\label{fig:1d-synthetic-data}
\end{figure}

We generate two one-dimensional (1D) and a three-dimensional (3D) synthetic data sets:
Each synthetic data set includes 1,000 sequences, with an average length of 150 events per sequence. 
Two 1D (time-only) data sets are simulated by a self-exciting process and a self-correcting process, respectively, using the thinning algorithm (Algorithm~\ref{alg:thinning-algorithm}).
The 3D (time and space) data set is generated by a randomly initialized \texttt{CEG+KDE} using Algorithm~\ref{alg:CEG-data-generation}. 

Three quantitative metrics are adopted for the evaluations on synthetic data sets: mean relative error (MRE) of the estimated conditional intensity and PDF compared to the ground truth on one randomly selected testing sequence, and the model log-likelihood on the entire testing set.
Note that we obtain these metrics for our model and other baselines in different ways. For our model, we estimate the conditional PDF using generated samples, based on which we can calculate the intensity function and model likelihood.
For the baselines, we need first to estimate the conditional intensity function based on their models and then compute the corresponding conditional PDF and log-likelihood by numerical approximations using (\ref{eq:cond-prob}) and (\ref{eq:log-likelihood}), respectively.


\begin{figure}[!t]
    \centering
    \begin{subfigure}[h]{.24\linewidth}
        {\includegraphics[width=\linewidth]{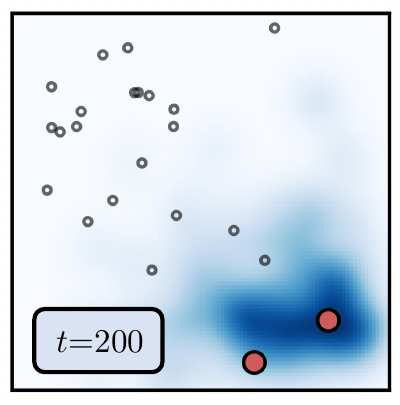}}
        {\includegraphics[width=\linewidth]{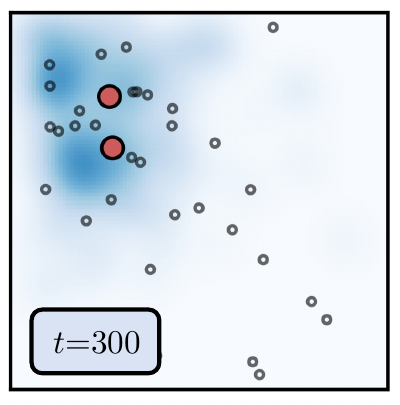}}
        {\includegraphics[width=\linewidth]{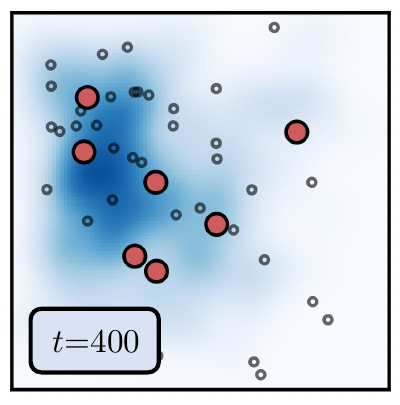}}
        \caption{Ground truth}
    \end{subfigure}
    \begin{subfigure}[h]{.24\linewidth}
        {\includegraphics[width=\linewidth]{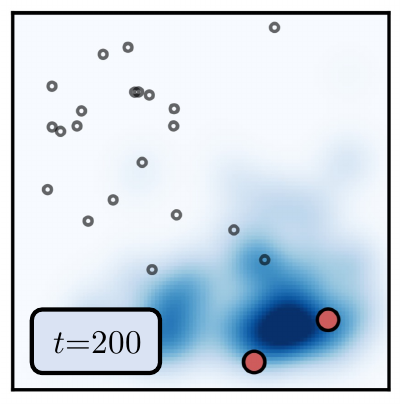}}
        {\includegraphics[width=\linewidth]{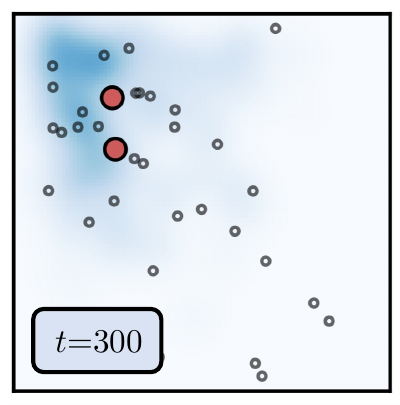}}
        {\includegraphics[width=\linewidth]{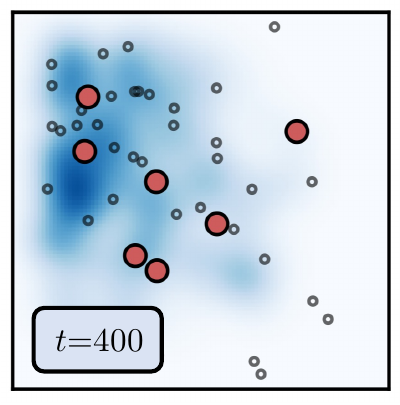}}
        \caption{\texttt{CEG+CDDM}}
    \end{subfigure}
    \begin{subfigure}[h]{.24\linewidth}
        {\includegraphics[width=\linewidth]{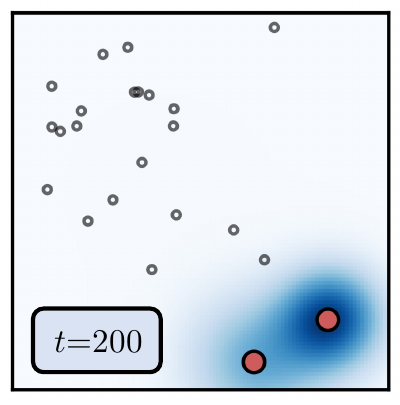}}
        {\includegraphics[width=\linewidth]{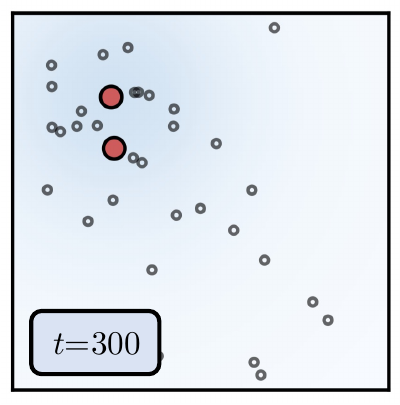}}
        {\includegraphics[width=\linewidth]{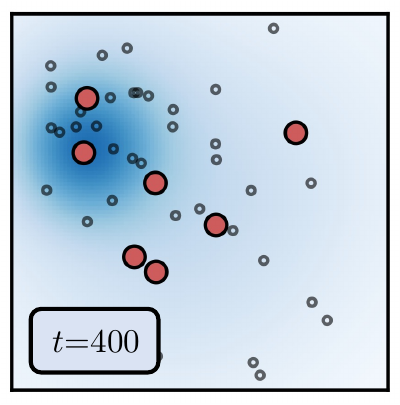}}
        \caption{\texttt{ETAS}}
    \end{subfigure}
    \begin{subfigure}[h]{.24\linewidth}
        {\includegraphics[width=\linewidth]{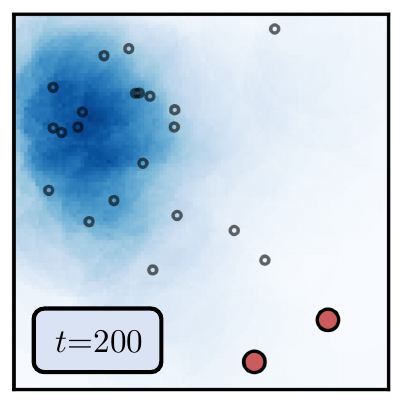}}
        {\includegraphics[width=\linewidth]{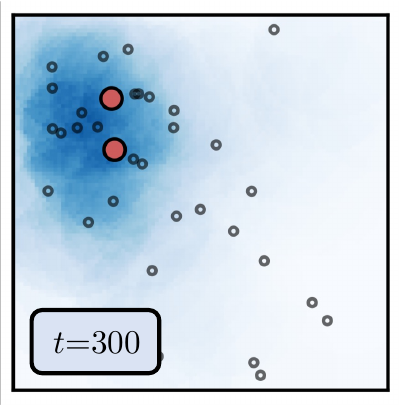}}
        {\includegraphics[width=\linewidth]{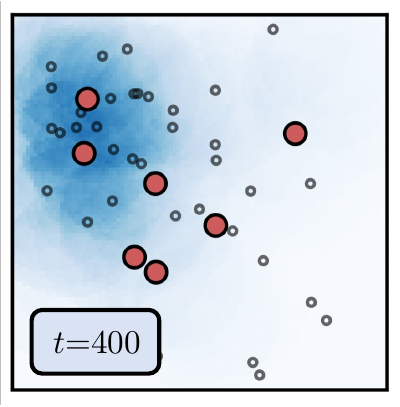}}
        \caption{\texttt{DNSK}}
    \end{subfigure}
\caption{Snapshots of out-of-sample estimation of the conditional PDFs for a three-dimensional (time and space) event sequence, arranged in chronological order from top to bottom. Darker shades indicate higher conditional PDF values. The red dots represent newly observed events, and the circles represent historical events.}
\label{fig:3d-synthetic-data}
\vspace{-0.05in}
\end{figure}

Figure~\ref{fig:1d-synthetic-data} shows the estimated conditional PDFs and intensities, as well as their corresponding ground truth of one testing sequence generated by the 1D self-exciting and self-correcting process, respectively.
All the variants of our generative model enjoy superior performance compared to baselines in accurately recovering the true conditional PDFs and intensities for both data sets. 
Similar results on 3D synthetic data are visualized in Figure~\ref{fig:3d-synthetic-data}, where each row displays four snapshots of estimated conditional PDFs for a testing sequence. Our model's estimated PDFs closely match the ground truth and capture the complex spatio-temporal point patterns, while \texttt{DNSK} and \texttt{ETAS} fails to capture the heterogeneous triggering effects between events, indicating limited practical representational power.

Table~\ref{tab:synthetic-data-results} presents quantitative results on 1D and 3D data sets, including the log-likelihood per testing event and the mean relative error (MRE) of the recovered conditional density and intensity (MRE is inapplicable on earthquake data for we have no ground truth). These results demonstrate the consistent comparable or superior performance of \texttt{CEG} against other methods across all scenarios. We note a performance disparity among different variants of \texttt{CEG}, which is introduced by the nature of different generators. 
The \texttt{CEG+KDE} has the best scores on 1D data sets because the KDE can provide an accurate estimation of one-dimensional data distribution with enough candidate points sampled (Appendix~\ref{app:non-param-learning}). However, it suffers from drastically increasing estimation error as the data dimension increases (\eg, seeing results on 3D data sets) and even becomes intractable in high-dimensional space. In these cases, using generative model architectures can be a better choice.
The performance of CVAE is limited on data with multimodal distribution due to its use of a simple isotropic Gaussian as the prior in the latent space \citep{lavda2019improving} (\eg, limited performance on 3D synthetic and earthquake data that both exhibit complex and multimodal event distributions (Figure~\ref{fig:3d-synthetic-data}/\ref{fig:real-generated-data}/\ref{fig:real-exps})). On the contrary, By leveraging the generative power of CDDM for multimodal distributions, we overcome this challenge and significantly improve the model's fit to data in multi- and high-dimensional spaces.


\subsection{Semi-synthetic data with image marks}

\begin{figure*}[!t]
    \centering
    \begin{subfigure}[h]{.32\linewidth}
    {\includegraphics[width=1\linewidth]{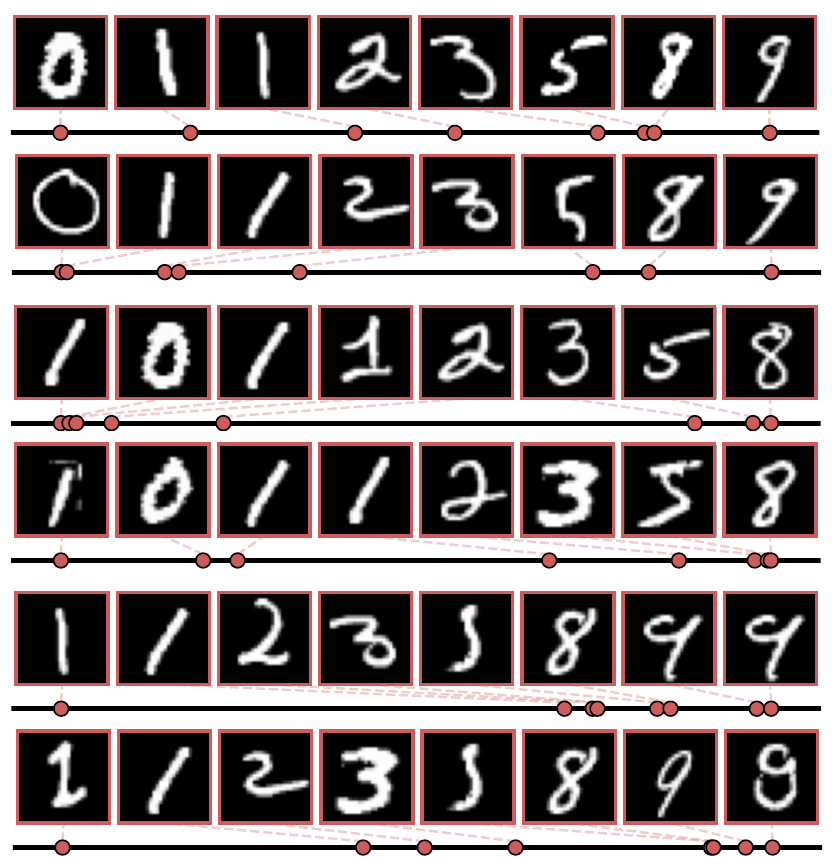}}
    \end{subfigure}
    \begin{subfigure}[h]{.32\linewidth}
    {\includegraphics[width=1\linewidth]{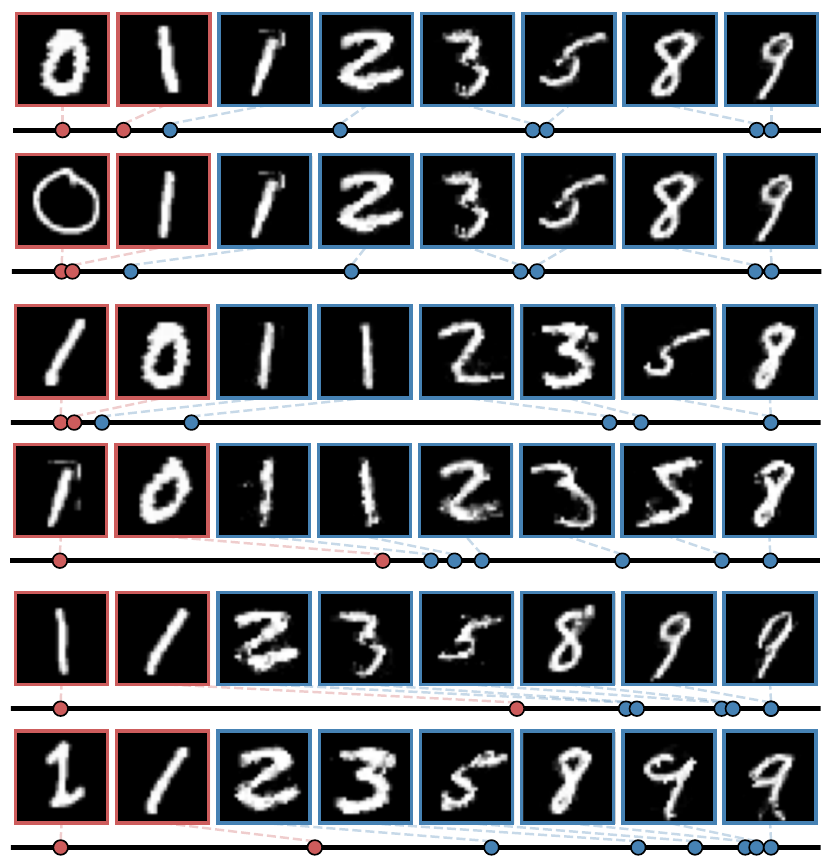}}
    \end{subfigure}
    \begin{subfigure}[h]{.32\linewidth}
    {\includegraphics[width=1\linewidth]{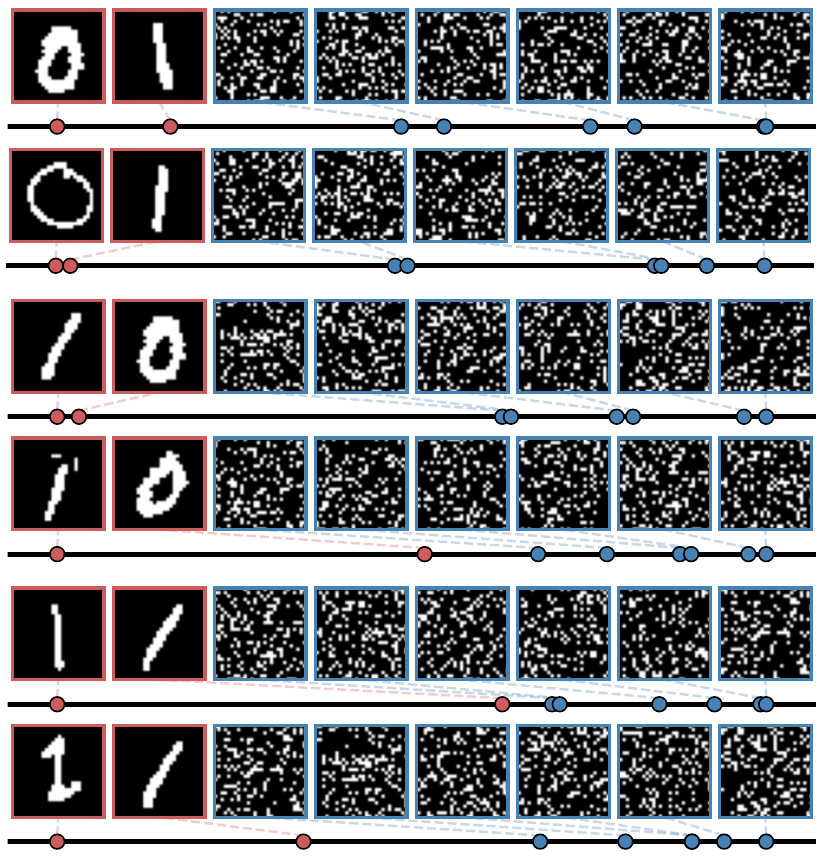}}
    \end{subfigure}

    \vspace{0.05in}
    \begin{subfigure}[h]{.32\linewidth}
    {\includegraphics[width=1\linewidth]{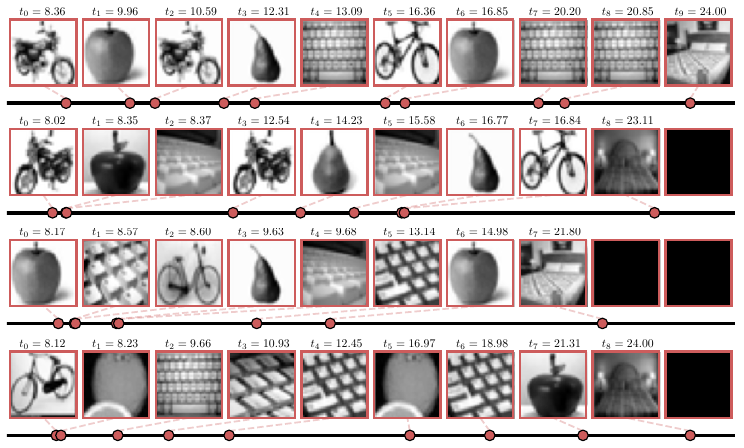}}
    \caption{True series}
    \end{subfigure}
    \begin{subfigure}[h]{.32\linewidth}
    {\includegraphics[width=1\linewidth]{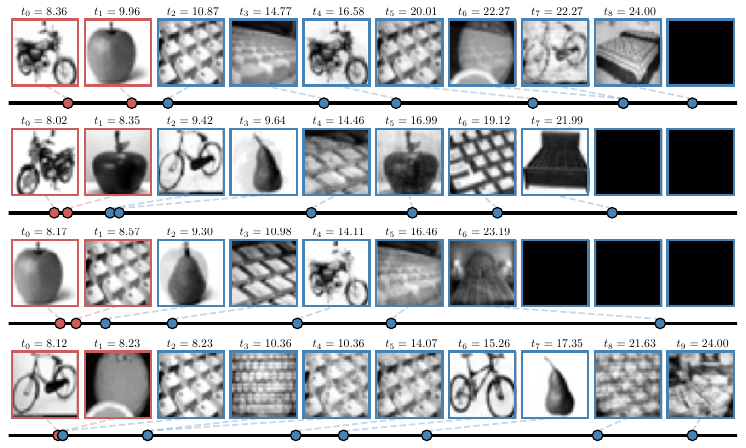}}
    \caption{\texttt{CEG+CDDM} generated series}
    \end{subfigure}
    \begin{subfigure}[h]{.32\linewidth}
    {\includegraphics[width=1\linewidth]{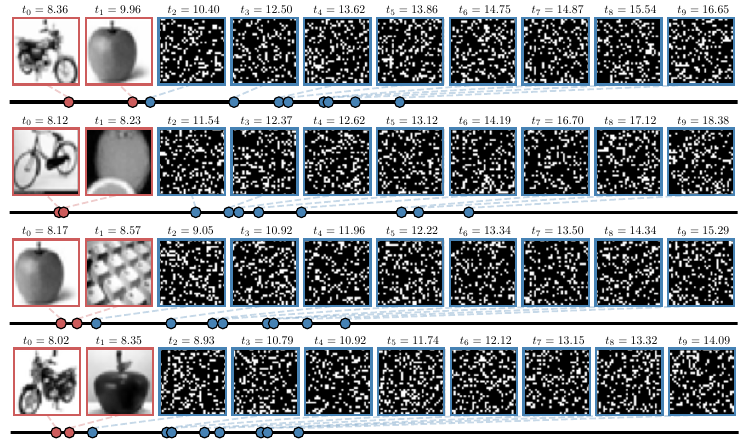}}
    \caption{\texttt{DNSK} generated series}
    \end{subfigure}
\caption{Generated \textsf{T-MNIST} (first row) and \textsf{T-CIFAR} (second row) series using \texttt{CEG+CDDM} and a neural point process baseline \texttt{DNSK}, with true sequences displayed on the left. Each series is generated (blue boxes) given the first two true events (red boxes). Dots on the horizontal lines represent the event times, with dashed lines indicating the association between time and mark. }
\label{fig:mnist-cifar-synthetic-data}
\vspace{-.05in}
\end{figure*}


We test our model's capability of generating complex high-dimensional marked events on two semi-synthetic data, including time-stamped MNIST (\textsf{T-MNIST}) and CIFAR-100 (\textsf{T-CIFAR}). 
In these data sets, both the mark (the image category) and the timestamp are generated through a marked point process. 
Images from MNIST and CIFAR-100 are subsequently chosen at random based on these marks, acting as a high-dimensional representation of the original image category. 
It's important to note that during the training phase, categorical marks are excluded, retaining only the high-dimensional images for model learning. Details of data generation are in Appendix~\ref{append:additional-results}.

\vspace{-0.05in}
\begin{enumerate}
    \item \textsf{T-MNIST}: For each sequence in the data, the actual digit in the succeeding image is the aggregate of the digits in the two preceding marks.
    The initial two digits are randomly selected from $0$ and $1$. The digits in the marks must be less than nine. The hand-written image for each mark is then chosen from the corresponding subset of MNIST according to the digit. 
    The time for the entire MNIST series conforms to a Hawkes process with an exponentially decaying kernel.
    \item \textsf{T-CIFAR}: The data contains event series that depict a typical day in the life of a graduate student, spanning from 8:00 to 24:00. The marks are sampled from four categories: outdoor exercises, food ingestion, working, and sleeping. 
    Depending on the most recent activity, the subsequent one is determined by a transition probability matrix. Images are selected from the respective categories to symbolize each activity. The time for these activities follows a self-correcting process.
    \vspace{-0.02in}
\end{enumerate}

\vspace{-0.1in}
\paragraph{Evaluation of the high-dimensional data sets} Since evaluating the log-likelihood for event series with high-dimensional marks is infeasible for our model (the number of samples needed to estimate the density at one data point is impractically large), we evaluate the model performance according to (i) the quality of the generated event time sequences, (ii) the quality of the generated image marks, and (iii) the transition dynamics of the series. For (i), we use the \textit{predictive score} (Pred.) and the \textit{discriminative score} (Disc.) proposed in \citep{yoon2019time} for measuring the fidelity and diversity of the synthesized time-series samples (See more details about these two metrics in Appendix~\ref{append:additional-results}). For (ii), we do not include the FID \citep{heusel2017gans} or IS \citep{salimans2016improved} because we are not focusing on the image generation problem, and the quality difference of the generated image marks from different models can be easily distinguished. Here, we compared our model \texttt{CEG+CDDM} with \texttt{DNSK}, since other baselines cannot handle high-dimensional event marks.

\vspace{0.05in}
Figure~\ref{fig:mnist-cifar-synthetic-data} presents the true \textsf{T-MNIST} series alongside the series generated by \texttt{CEG+CDDM} and \texttt{DNSK} given the first two events. Our model not only generates high-dimensional marks that resemble true images, but also successfully captures the underlying data dynamics, such as the clustering of events and the transition pattern of image marks. On the contrary, the \texttt{DNSK} only learns the temporal effects of events and struggles to estimate the conditional intensity for high-dimensional marks. Besides, the grainy images generated by \texttt{DNSK} showcase the challenge of simulating credible high-dimensional content using the thinning algorithm, limiting the usefulness of previous approaches in such an application.
This is because the real data points, being sparsely scattered in the high-dimensional mark space, make it challenging for the thinning algorithm to sample candidate points that align closely with them.
Metrics in Table~\ref{tab:quantitative-hd-data-results} emphasize that our generative framework can accurately predict the timing of marks’ occurrence, a task that extends beyond more difficult than mere image or text generation.



\begin{table}[!t]
  \caption{Quantitative evaluation of the generated event time sequences with high-dimensional marks.}
  \vspace{-.1in}
  \centering
  \resizebox{1.\linewidth}{!}{
  \begin{threeparttable}
  \begin{tabular}{ccccccc}
    \toprule
    \toprule
    & \multicolumn{2}{c}{\bf \textsf{T-MNIST}} & \multicolumn{2}{c}{\bf \textsf{T-CIFAR}} & \multicolumn{2}{c}{\bf Textual crime} \\
    \cmidrule(lr){2-3} \cmidrule(lr){4-5} \cmidrule(lr){6-7}
    Model & Disc.($\downarrow$) & Pred.($\downarrow$) & Disc. & Pred. & Disc. & Pred. \\
    \midrule
    \texttt{DNSK} & $0.388~(0.021)$ & $\textbf{0.003}~(0.001)$ & $0.375~(0.050)$ & $0.031~(0.002)$ & $0.500~(0.000)$ & $0.010~(0.008)$ \\
    \texttt{CEG+CVAE} & $0.097~(0.037)$ & $\textbf{0.003}~(0.002)$ & $0.189~(0.023)$ & $0.010~(0.002)$ & $0.102~(0.016)$ & $0.003~(0.001)$ \\
    \texttt{CEG+CDDM} & $\textbf{0.063}~(0.034)$ & $\textbf{0.003}~(<0.001)$ & $\textbf{0.168}~(0.027)$ & $\textbf{0.006}~(<0.001)$ & $\textbf{0.071}~(0.014)$ & $\textbf{0.002}~(<0.001)$ \\
    \bottomrule
    \bottomrule
  \end{tabular}
  \begin{tablenotes}
  \item *Numbers in parentheses present standard error for three independent runs.
  \end{tablenotes}
  \end{threeparttable}
  }
  \label{tab:quantitative-hd-data-results}
\end{table}

Similar results are shown in Figure~\ref{fig:mnist-cifar-synthetic-data} and Table~\ref{tab:quantitative-hd-data-results} on \textsf{T-CIFAR} data, where \texttt{CEG+CDDM} can simulate high-quality daily activities with high-dimensional content at appropriate times. However, \texttt{DNSK} fails to extract any meaningful patterns from data since intensity-based modeling or data generation becomes ineffectual in high-dimensional mark space.

\subsection{Real data}

In our real data results, our model also demonstrates superior efficacy in generating high-quality multi- or high-dimensional event sequences, using earthquake and crime data sets. 

\vspace{-0.1in}
\paragraph{Northern California earthquake catalog} We test our method using the Northern California Earthquake Data \citep{NCEDC2014}, which contains detailed information on the timing and location of earthquakes that occurred in central and northern California from 1978 to 2018, totaling 5,984 records with magnitude greater than 3.5. We divided the data into several sequences by month.
In comparison to other baseline methods that can only handle 1D event data, we primarily evaluated our model against \texttt{DNSK} and \texttt{ETAS}, training each model using 80\% of the dataset and testing them on the rest. To demonstrate the effectiveness of our method on the real data, we assess the quality of the generated sequences by each model. Our model's generation process for new sequences can be efficiently carried out using Algorithm~\ref{alg:CEG-data-generation}, whereas both \texttt{DNSK} and \texttt{ETAS} requires the use of a thinning algorithm (Algorithm~\ref{alg:thinning-algorithm}) for simulation. We also compared the estimated conditional probability density functions (PDFs) of real sequences by each model in Appendix~\ref{append:additional-results}.

\begin{figure}[!t]
    \centering
    \begin{subfigure}[h]{.235\linewidth}
        \fcolorbox{black}{white}{\includegraphics[width=\linewidth]{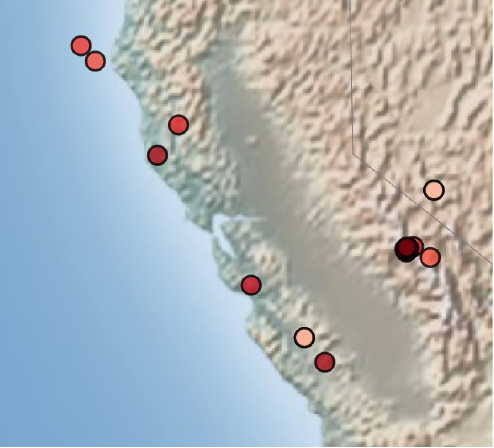}}
        \fcolorbox{black}{white}{\includegraphics[width=\linewidth]{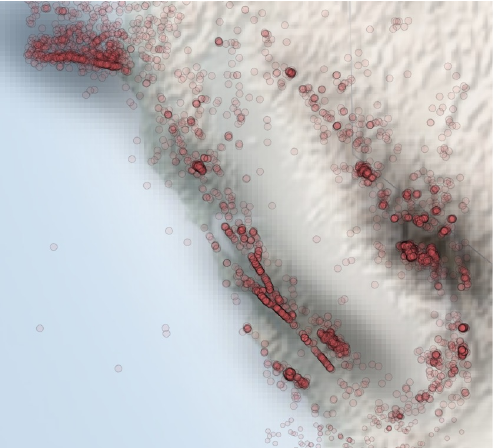}}
        \caption{Real}
    \end{subfigure}
    \begin{subfigure}[h]{.235\linewidth}
        \fcolorbox{black}{white}{\includegraphics[width=\linewidth]{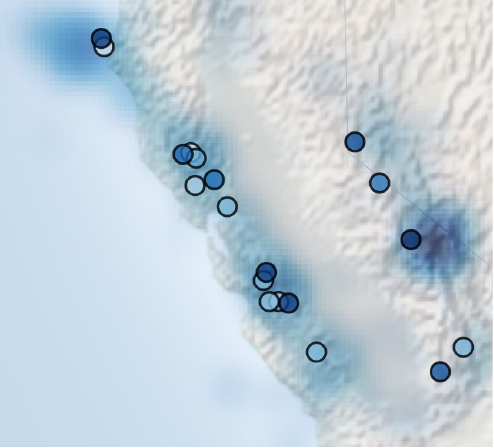}}
        \fcolorbox{black}{white}{\includegraphics[width=\linewidth]{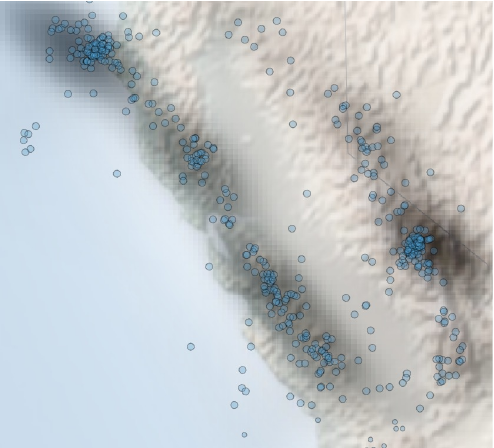}}
        \caption{\texttt{CEG+CDDM}}
    \end{subfigure}
    \begin{subfigure}[h]{.235\linewidth}
        \fcolorbox{black}{white}{\includegraphics[width=\linewidth]{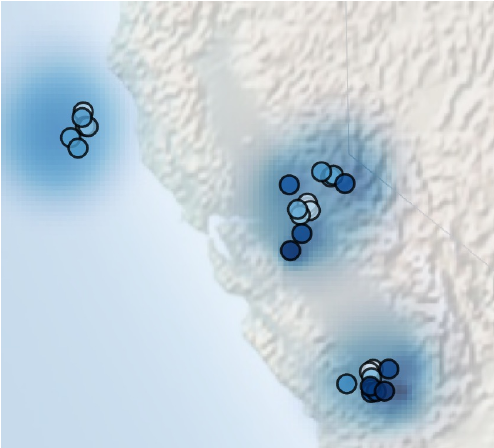}}
        \fcolorbox{black}{white}{\includegraphics[width=\linewidth]{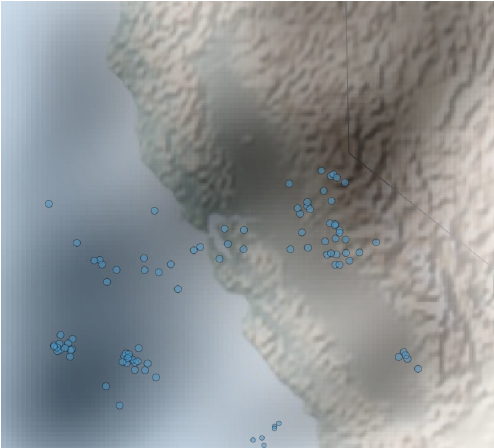}}
        \caption{\texttt{ETAS}}
    \end{subfigure}
    \begin{subfigure}[h]{.235\linewidth}
        \fcolorbox{black}{white}{\includegraphics[width=\linewidth]{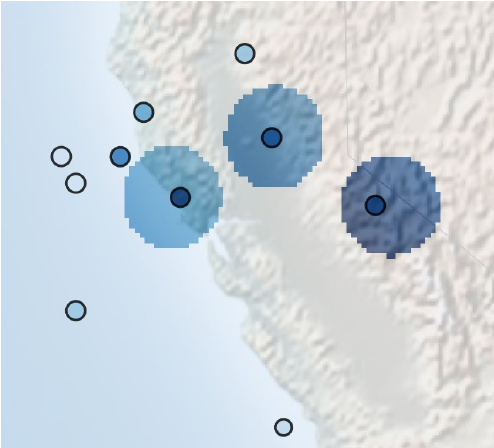}}
        \fcolorbox{black}{white}{\includegraphics[width=\linewidth]{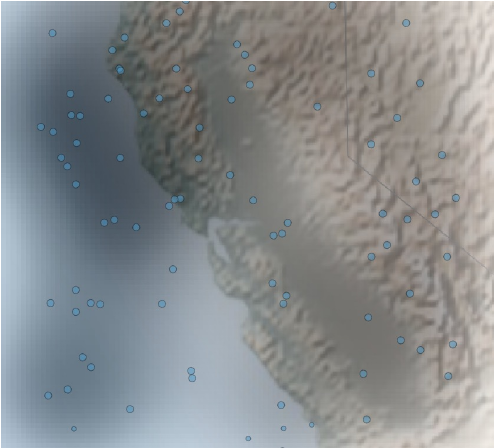}}
        \caption{\texttt{DNSK}}
    \end{subfigure}
\caption{Real and generated earthquake sequences. The first row displays a single sequence, either real or generated, with the color depth of the dots reflecting the occurrence time of each event (darker colors for more recent events). The shaded blue areas represent the estimated conditional PDFs. Pictures in the second row overlay 1,000 real or generated events together. The grey shaded area indicates the event density estimated by KDE, which can be interpreted as the background rate of events.}
\label{fig:real-generated-data}
\vspace{-.05in}
\end{figure}

\begin{figure}[!t]
    \begin{subfigure}[h]{\linewidth}
        \centering
        \fcolorbox{black}{white}{\includegraphics[width=.235\linewidth]{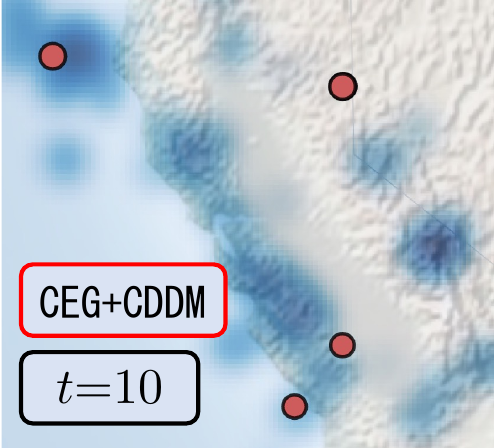}}
        \fcolorbox{black}{white}{\includegraphics[width=.235\linewidth]{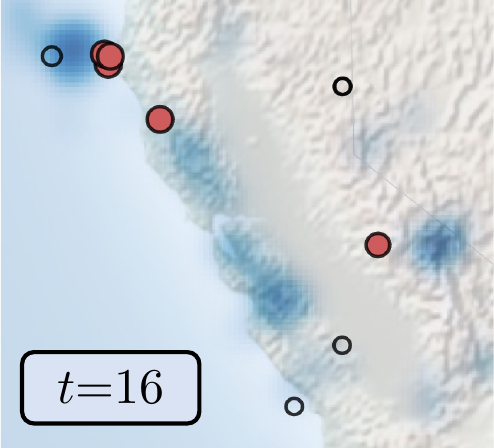}}
        \fcolorbox{black}{white}{\includegraphics[width=.235\linewidth]{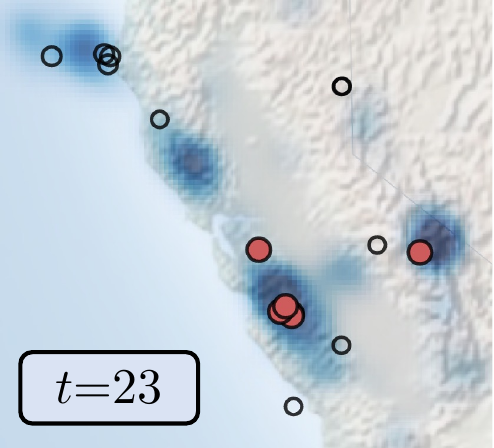}}
        \fcolorbox{black}{white}{\includegraphics[width=.235\linewidth]{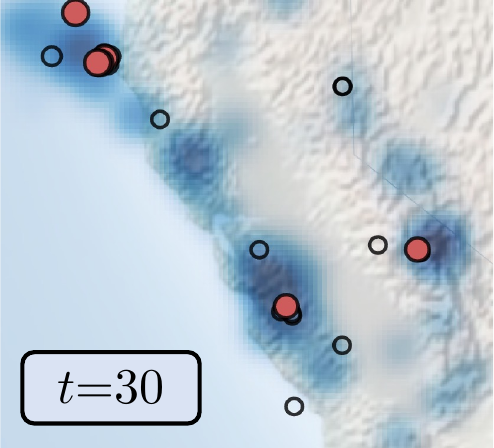}}
    \end{subfigure}
    \begin{subfigure}[h]{\linewidth}
        \centering
        \fcolorbox{black}{white}{\includegraphics[width=.235\linewidth]{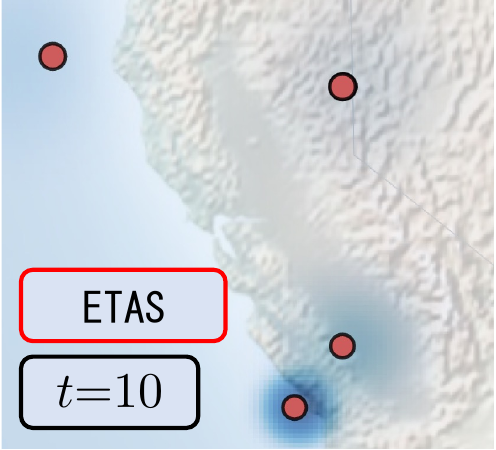}}
        \fcolorbox{black}{white}{\includegraphics[width=.235\linewidth]{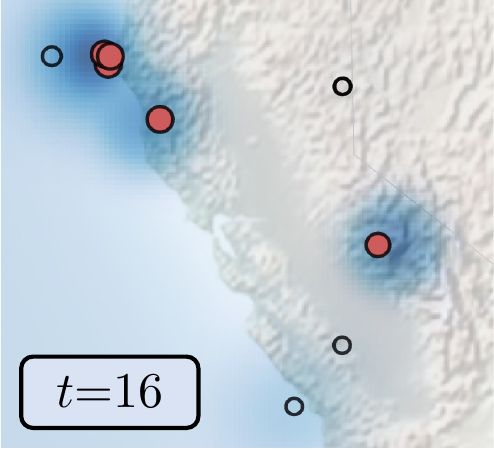}}
        \fcolorbox{black}{white}{\includegraphics[width=.235\linewidth]{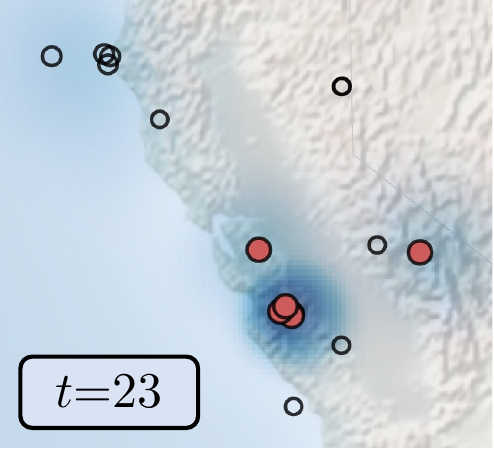}}
        \fcolorbox{black}{white}{\includegraphics[width=.235\linewidth]{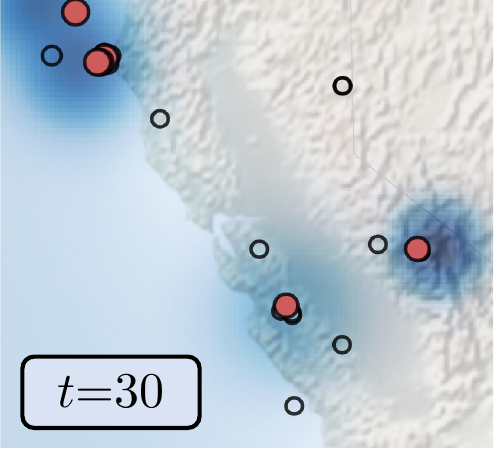}}
    \end{subfigure}
    \begin{subfigure}[h]{\linewidth}
        \centering
        \fcolorbox{black}{white}{\includegraphics[width=.235\linewidth]{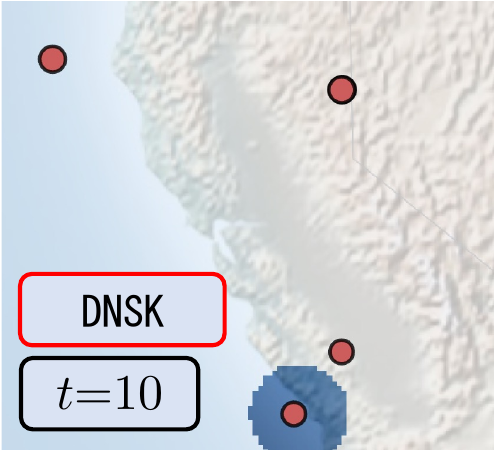}}
        \fcolorbox{black}{white}{\includegraphics[width=.235\linewidth]{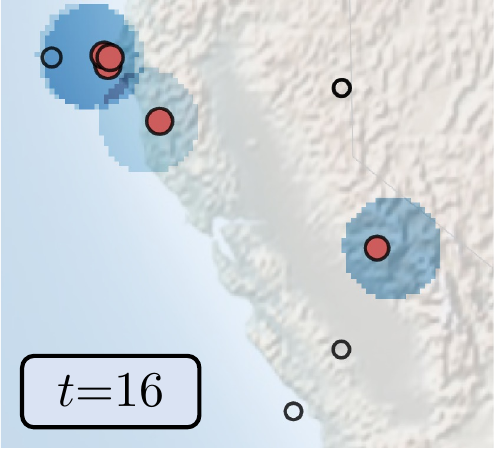}}
        \fcolorbox{black}{white}{\includegraphics[width=.235\linewidth]{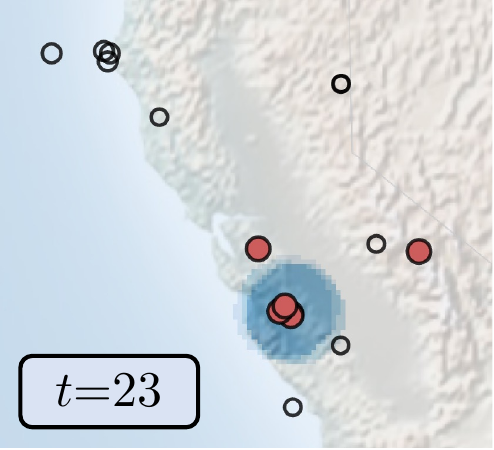}}
        \fcolorbox{black}{white}{\includegraphics[width=.235\linewidth]{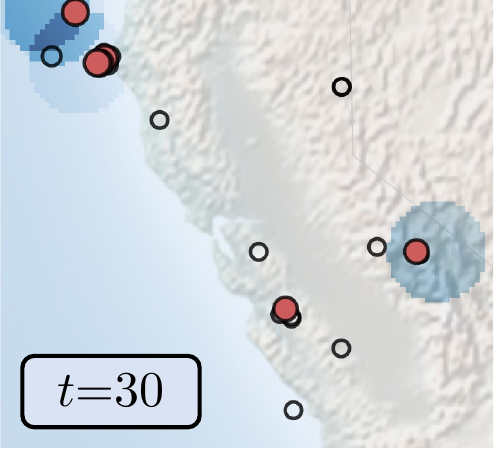}}
    \end{subfigure}
\caption{Estimated conditional PDFs (shaded areas) of an actual earthquake sequence, with darker shades indicating higher values. Each row contains four sub-figures, arranged in chronological order from left to right, showing snapshots of the estimated conditional PDFs by each model. The red dots represent the most recently observed events, and the circles represent historical events.}
\vspace{-.05in}
\label{fig:real-exps}
\end{figure}

We compare models' generative ability in Figure~\ref{fig:real-generated-data}. The first panel shows a single event series randomly selected from the data, while the rest panels in the first row exhibit event series generated by each model. The quality of the generated earthquake sequence by our method is markedly superior to that generated by \texttt{DNSK} and \texttt{ETAS}.
We also simulate multiple sequences and visualize the spatial distribution of generated earthquakes in the second row. The shaded area reflects the spatial density of earthquakes obtained by KDE and represents the ``background rate'' over space. It is evident that \texttt{CEG+CDDM} is successful in capturing the underlying earthquake distribution, while the two STPP baselines are unable to do so.

Figure~\ref{fig:real-exps} visualizes the conditional PDF estimated by \texttt{CEG+CDDM}, \texttt{DNSK}, and \texttt{ETAS} for an actual earthquake series in the testing set. Our model's finding of a heightened probability of seismic activity along the San Andreas fault, coupled with a diminished likelihood in the basin, aligns with current understandings of the earthquake mechanics in Northern California \citep{wallace1990san}. 
However, both \texttt{DNSK} and \texttt{ETAS} fail to extract this geographical feature from the data, suggesting a uniform impact of observed earthquakes on the surroundings.


\vspace{-0.1in}
\paragraph{Atlanta crime reports with textual description}
We further assess our method using 911-calls-for-service data in Atlanta. 
The proprietary data set contains 4644 burglary incidents from 2016 to 2017,
detailing the time, location, and a comprehensive textual description of each incident.
Each textual description was transformed into a TF-IDF vector \citep{aizawa2003information}, from which the top 10 keywords with the most significant TF-IDF values were selected.
The location combined with the corresponding 10-dimensional TF-IDF vector is regarded as the mark of the incident.  
We fit \texttt{CEG+CDDM} and \texttt{DNSK} using the preprocessed data, subsequently generate crime event sequences, and then compare them with the real data.

\begin{figure*}[!t]
    \centering
    \begin{subfigure}[h]{.19\linewidth}
        {\includegraphics[width=\linewidth]{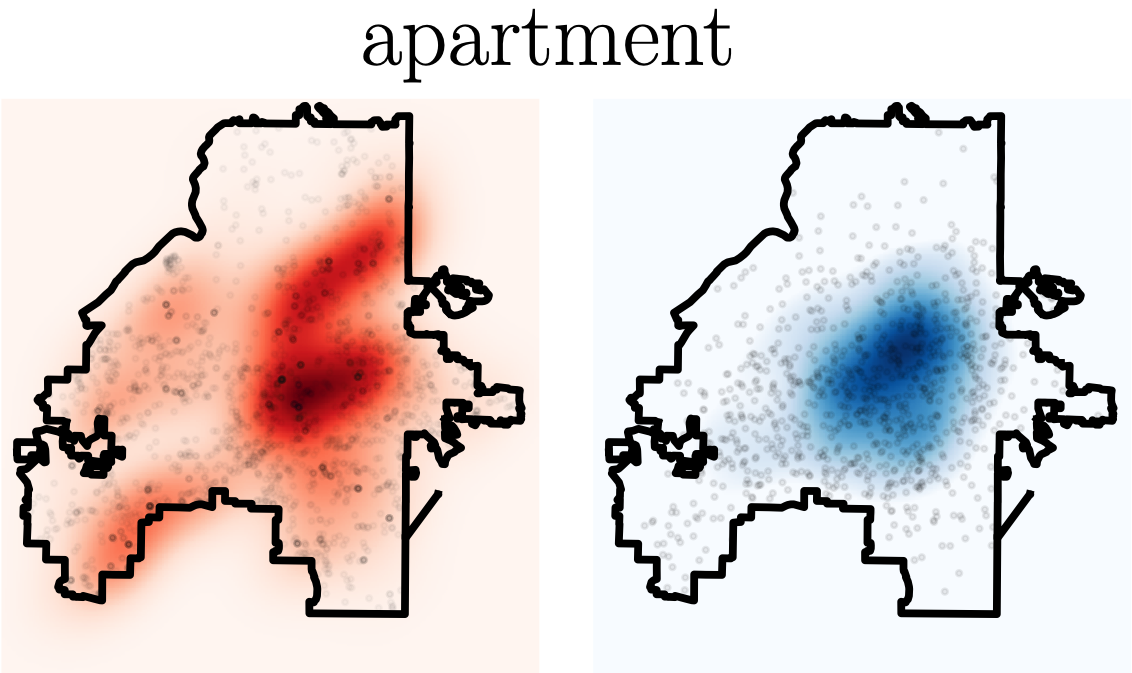}}
        {\includegraphics[width=\linewidth]{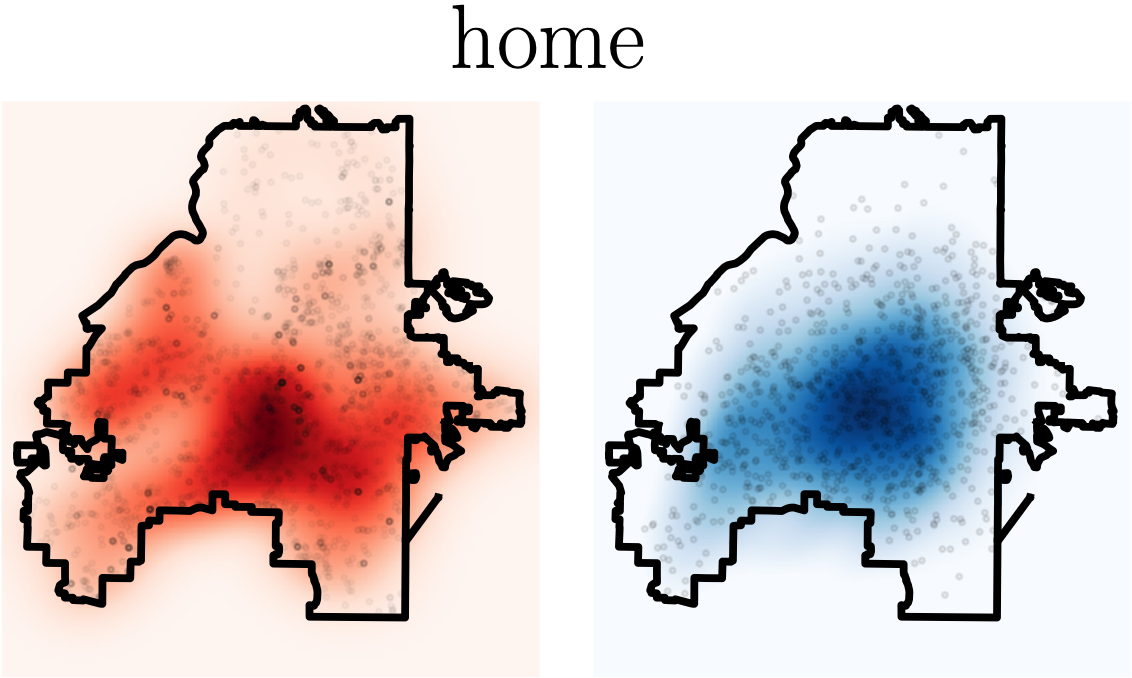}}
    \end{subfigure}
    \begin{subfigure}[h]{.19\linewidth}
        {\includegraphics[width=\linewidth]{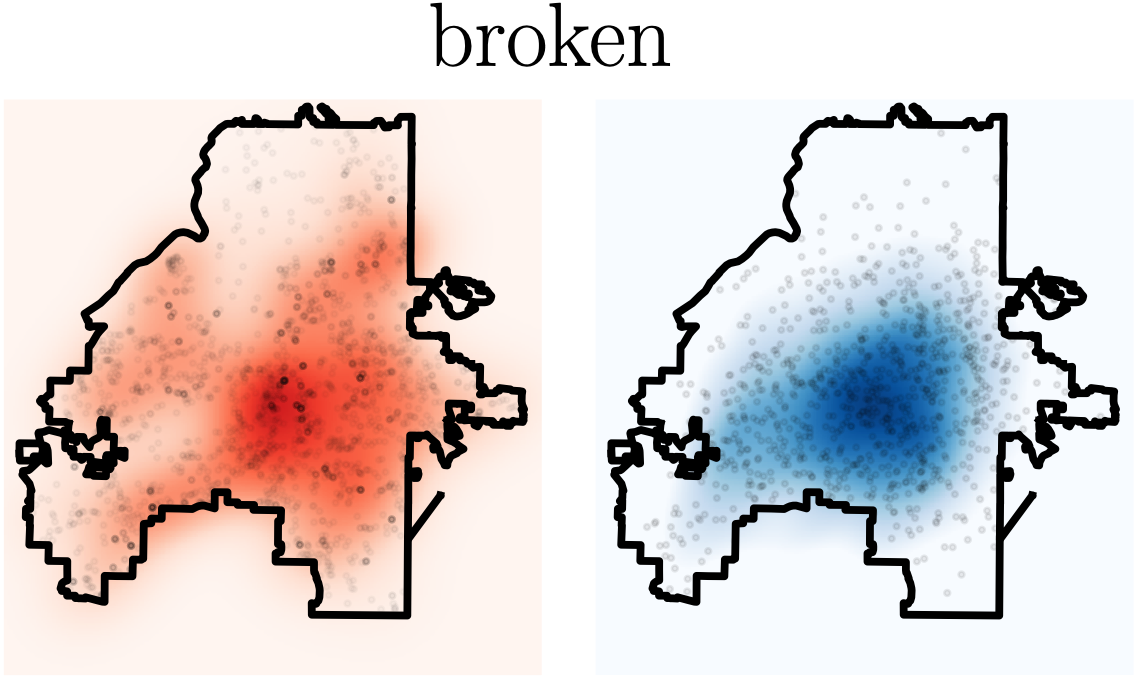}}
        {\includegraphics[width=\linewidth]{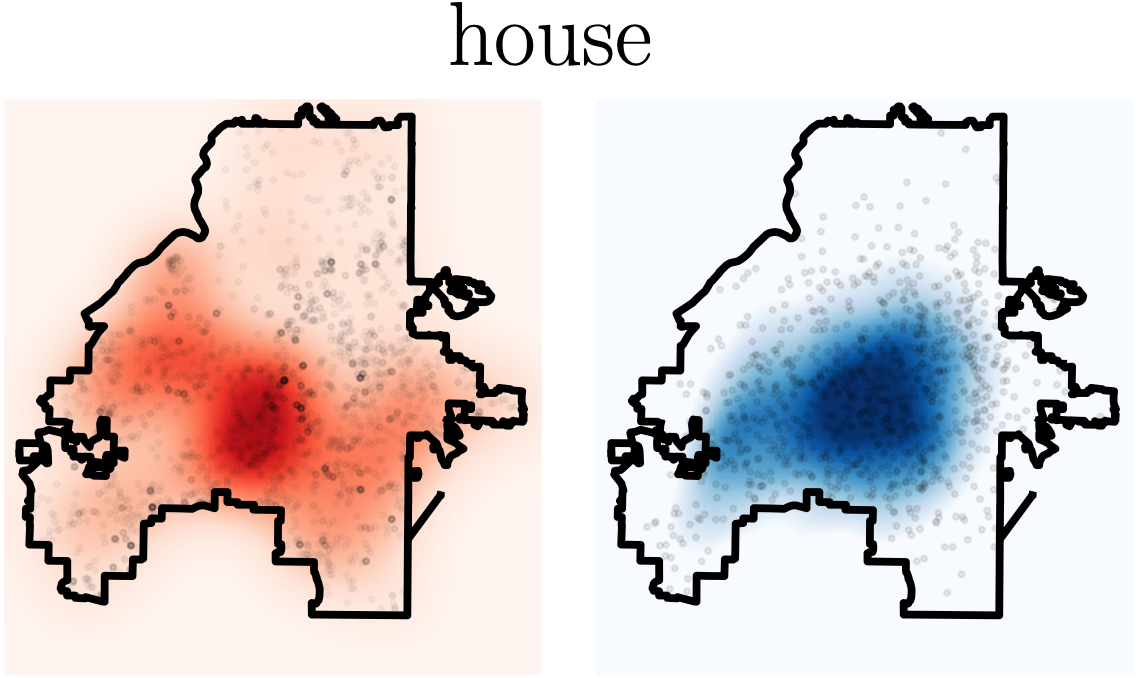}}
    \end{subfigure}
    \begin{subfigure}[h]{.19\linewidth}
        {\includegraphics[width=\linewidth]{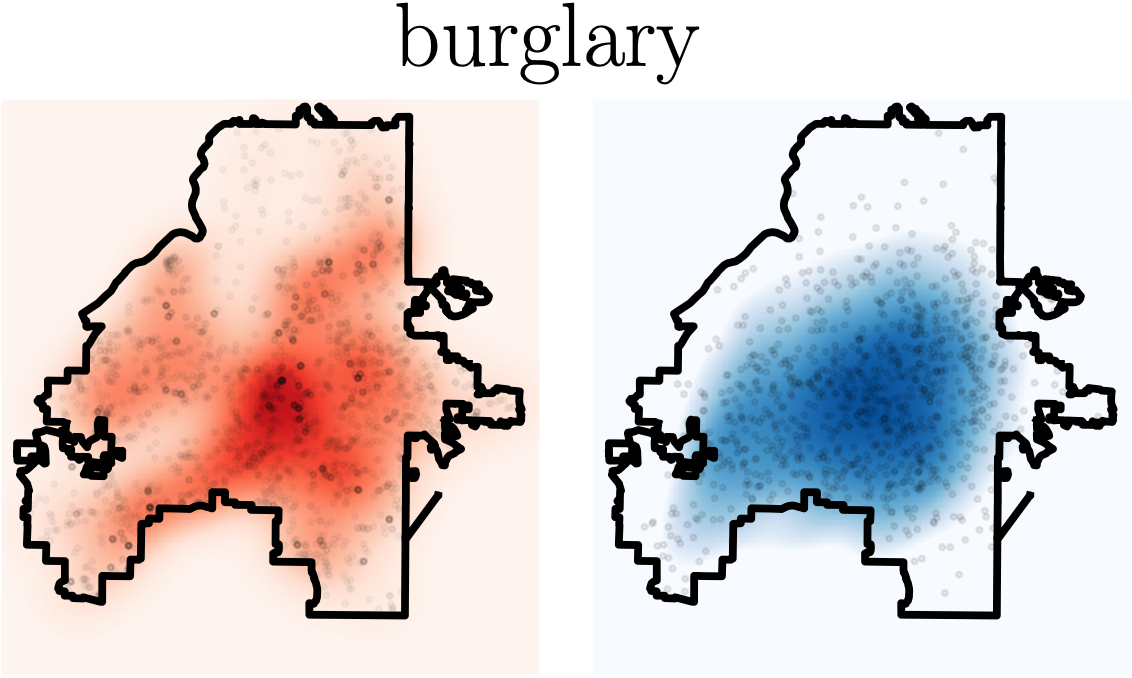}}
        {\includegraphics[width=\linewidth]{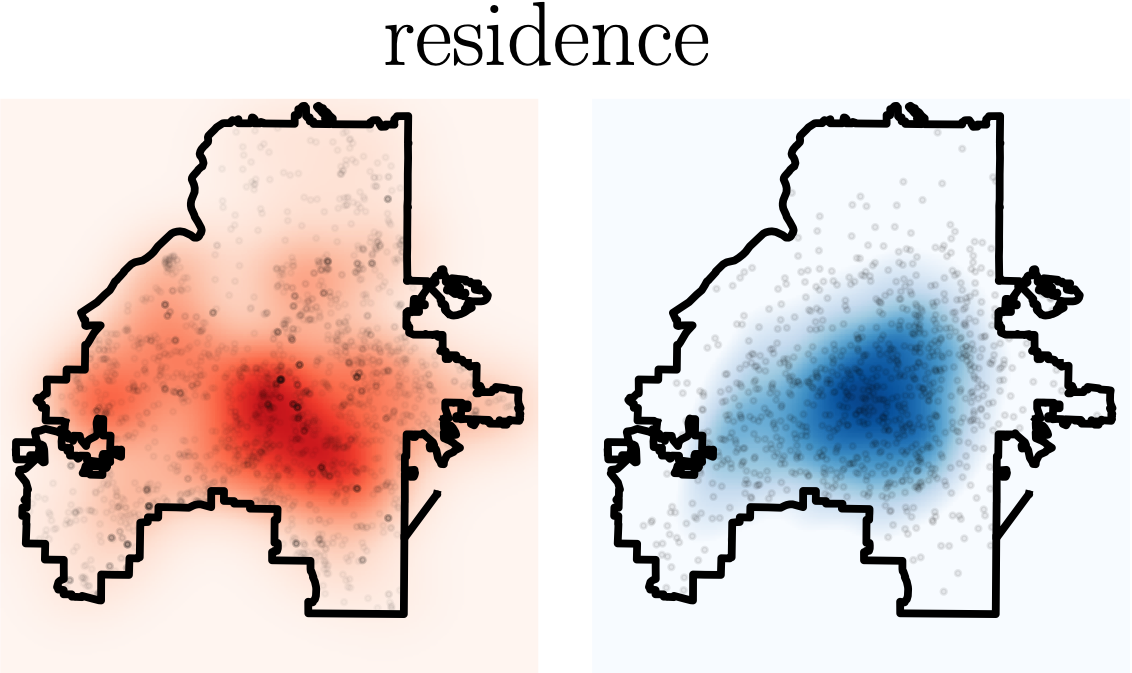}}
    \end{subfigure}
    \begin{subfigure}[h]{.19\linewidth}
        {\includegraphics[width=\linewidth]{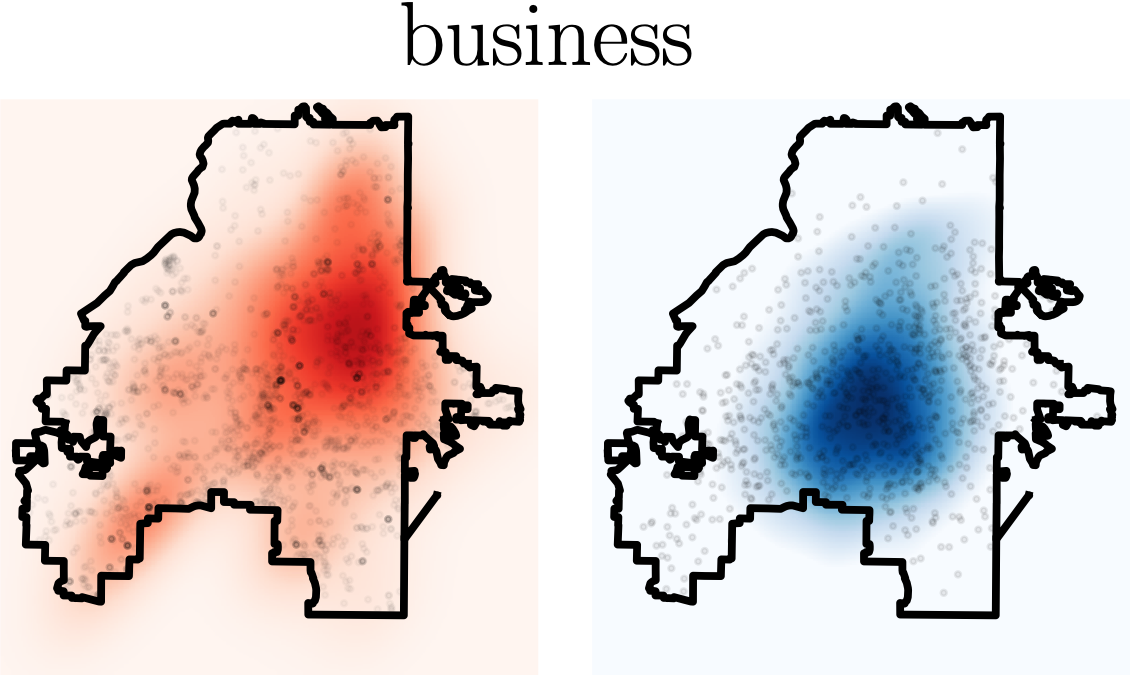}}
        {\includegraphics[width=\linewidth]{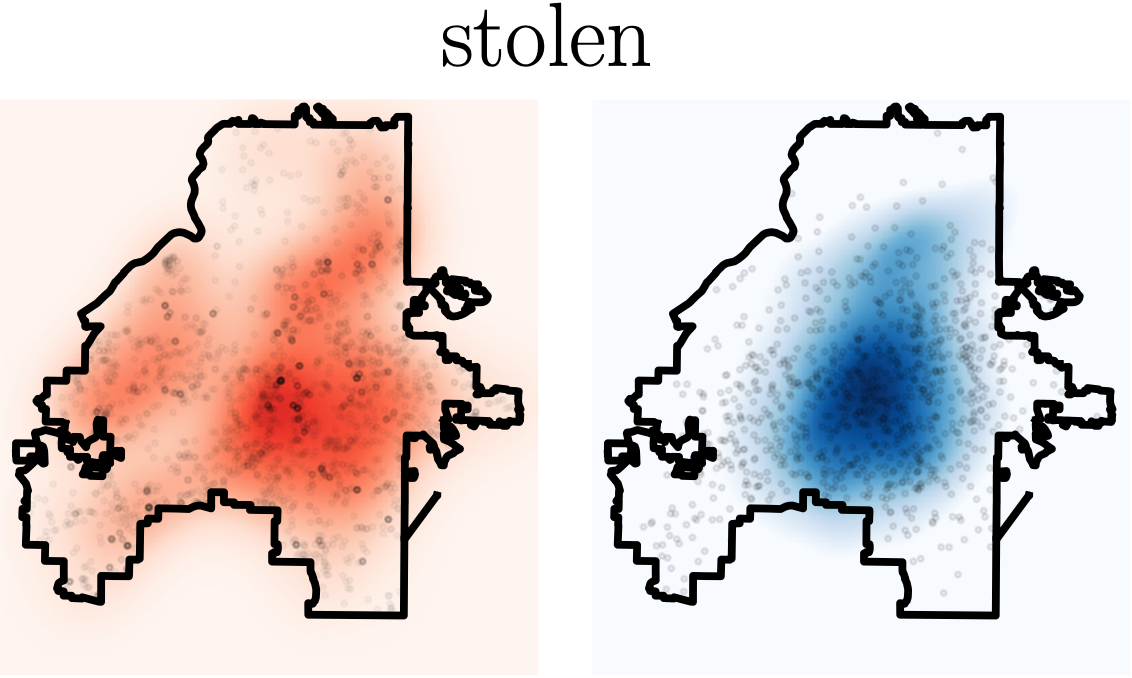}}
    \end{subfigure}
    \begin{subfigure}[h]{.19\linewidth}
        {\includegraphics[width=\linewidth]{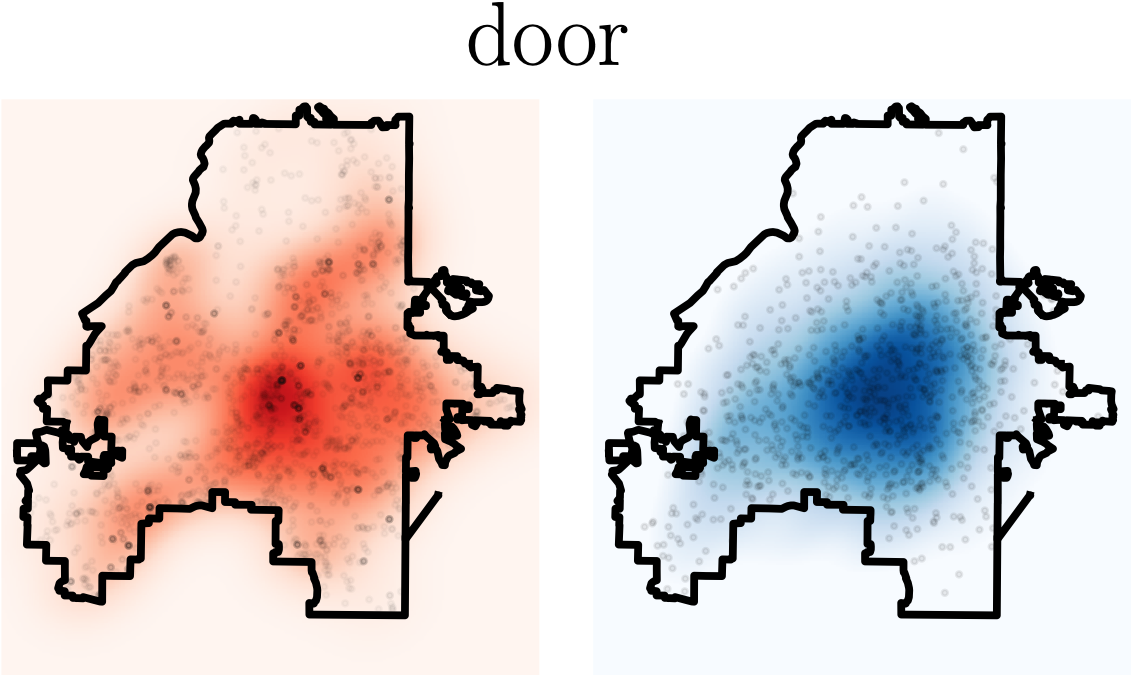}}
        {\includegraphics[width=\linewidth]{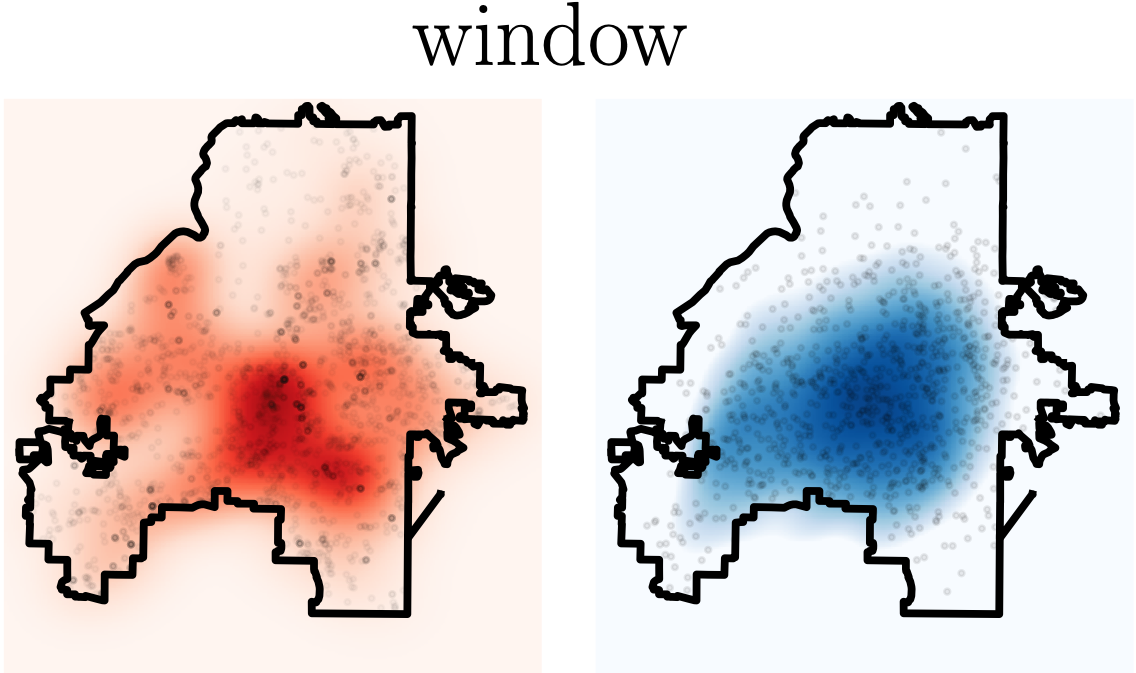}}
    \end{subfigure}
\caption{The spatial distributions of the TF-IDF values of 10 crime-related keywords. The heatmap in red and blue represent distributions of TF-IDF value of the keywords in the true and generated events, respectively. The black dots pinpoint the locations of the corresponding events.
}
\label{fig:police-text-generated-keywords}
\vspace{-.05in}
\end{figure*}

Figure~\ref{fig:police-text-generated-keywords} visualizes the spatial distributions of the true and the generated TF-IDF value of each keyword, respectively, signifying the heterogeneous crime patterns across the city. As we can observe, our model is capable of capturing such spatial heterogeneity for different keywords and simulating crime incidents that follow the underlying spatio-temporal-textual dynamics existing in criminological \emph{modus operandi} \citep{zhu2022spatiotemporal}. Meanwhile, quantitative measurements for the generated event time sequences are reported in Table~\ref{tab:quantitative-hd-data-results}, showing the capability of our model to learn and reproduce the temporal properties of the original crime reports. Note that our model is able to simultaneously capture the data dynamics in time, location, and mark space, despite the dimensionality difference in various modalities.

\subsection{Model efficiency comparison}

The thinning algorithm can suffer from low sampling efficiency for point process data because (i) it requires uniform sampling in the space $\mathcal{X}$ with an upper limit of the conditional intensity $\overline\lambda > \lambda(x),~\forall x$, while only a few candidates are retained in the end. (ii) Whether to reject one data point requires the evaluation of the conditional intensity over the stochastic event history. This doubly stochastic trait makes the entire thinning process costly when $\mathcal{X}$ is multi-dimensional, since it requires a drastically large number of candidate points and numerous evaluations of the conditional intensity function.

On the contrary, our model directly generates samples based on the learned event distribution, thus every generated point will be retained. 
Table~\ref{tab:generating-time-comparison} compares the costs for \texttt{ETAS}, \texttt{DNSK}, and \texttt{CEG+CDDM} to generate event series of length $100$. Note that our model requires a similar cost to generate different numbers of sequences. This is because \texttt{CEG+CDDM} can generate sequences in parallel, leveraging the benefits of the implementation of conditional generative models.

\begin{table}[!t]
  \caption{Time cost for model training and generating $5$ or $50$ event series of length $100$ using \texttt{ETAS}, \texttt{DNSK} and \texttt{CEG+CDDM}.}
  \vspace{-0.1in}
  \centering
  \resizebox{1.\linewidth}{!}{
  \begin{threeparttable}
    \begin{tabular}{cccccccccc}
    \toprule
    \toprule
    & \multicolumn{3}{c}{3D earthquake data} & \multicolumn{3}{c}{\textsf{T-MNIST}} & \multicolumn{3}{c}{\textsf{T-CIFAR}} \\
    \cmidrule(lr){2-4} \cmidrule(lr){5-7}  \cmidrule(lr){8-10}
    Model & Training & $5$ series & $50$ series & Training & $5$ series & $50$ series & Training & $5$ series & $50$ series \\
    \midrule
    \texttt{ETAS} & $0.742$ & $12.4$ & $118.6$ & / & / & / & / & / & / \\
    \texttt{DNSK} & $3.038$ & $20.1$ & $220.4$ & $9.150$ & $87.3$ & $745.6$ & $10.027$ & $274.0$ & $1381.9$ \\
    \texttt{CEG+CDDM}  & $0.869$ & $<0.1$ & $<0.1$ & $1.481$ & $0.6$ & $0.8$ & $1.532$ & $1.1$ & $1.2$ \\
    \bottomrule
    \bottomrule
  \end{tabular}
  \begin{tablenotes}
  \item *Unit for training time: second per epoch. Unit for inference cost: second.
  \end{tablenotes}
  \end{threeparttable}
  }
  \vspace{-0.05in}
  \label{tab:generating-time-comparison}
\end{table}


Our framework not only enjoys superior efficiency in sampling but also in model learning. As shown by the model training time in Table~\ref{tab:generating-time-comparison}, our generative framework improves the learning efficiency of the deep kernel-based approach \texttt{DNSK} by a large margin and achieves a comparable efficiency with the parametric model \texttt{ETAS} that only has three parameters to be estimated.


\vspace{-.0in}
\section{Discussions}

We introduce a novel framework that serves as a highly adaptable and efficient solution for modeling and generating temporal point process data with high-dimensional marks.
The proposed framework uses a conditional denoising diffusion model as the conditional event generator to explore the intricate multi-dimensional event space and generate subsequent events based on prior observations with exceptional efficiency. 
Numerical results demonstrate the superior performance of our model in capturing complex data distribution and generating high-quality samples against state-of-the-art methods, and its flexibility of being adapting to different real-world scenarios.
Finally, several promising areas for future research induced by our method can be foreseen. One possibility involves the incorporation of more advanced generative models to further facilitate the generation of high-dimensional marks. 
Moreover, the framework holds the potential to be adapted into a decision-making tool for sequential decision problems, such as in dynamic traffic management \citep{Liu2025} and real-time healthcare systems \citep{yu2021reinforcement}, where dynamic data and timely decisions are crucial.







\section*{Acknowledgement}

This research is partially supported by the NSF Grant CAIG-2425888.

\bibliographystyle{apalike}
\bibliography{main}

\newpage
\appendix

\setcounter{figure}{0} \renewcommand{\thefigure}{\thesection\arabic{figure}}
\setcounter{table}{0} \renewcommand{\thetable}{\thesection\arabic{table}}
\setcounter{equation}{0} \renewcommand{\theequation}{\thesection\arabic{equation}}

\section{Equivalent forms of log-likelihood}
\label{append:derivation-cond-prob}

Equivalence between \eqref{eq:log-likelihood} and \eqref{eq:sequence-likelihood} is drawn from the chain rule as
\[
    \begin{aligned}
        \ell(x_1, \dots, x_{N_T}) &= \log f(x_1, \dots, x_{N_T}) = \log \prod_{i=1}^{N_T}f(x_i|\mathcal{H}_{t_i}) \\
        &= \int_{\mathcal X}  \log \lambda(x| \mathcal{H}_{t(x)} ) d \mathbb N(x) -  \int_{\mathcal X}  \lambda(x| \mathcal{H}_{t(x)} ) dx.
    \end{aligned}
\]
The log-likelihood of $K$ observed event sequences in (\ref{eq:sequence-likelihood}) can be conveniently obtained with the counting measure $\mathbb{N}$ replaced by the counting measure $\mathbb{N}_k$ for the $k$-th sequence.

\section{Non-parametric learning}
\label{app:non-param-learning}

The self-tuned kernel with boundary correction for KDE is implemented as following:
\begin{enumerate}
    \item The bandwidth $\sigma$ tends to be small for those samples falling into event clusters and to be large for those isolated samples. 
    We dynamically determine the value of $\sigma$ for each sample $\widetilde{x}$ by computing the $k$-nearest neighbor ($k$NN) distance among other generated samples. 
    \item To correct the boundary bias, we use the kernel with reflection, defined as follows:
    \[
        \kappa(x - \widetilde{x}) = \upsilon^*(\Delta t - \Delta \widetilde{t} ) \cdot \upsilon(m - \widetilde{m}),
    \]
    where $\upsilon$ is an arbitrary kernel and $\upsilon^*(x - \widetilde{x}) = \upsilon(x - \widetilde{x}) + \upsilon(- x - \widetilde{x})$ is the same kernel with reflection boundary. 
    This allows for a more accurate estimation of the density near the boundary of the time domain without impacting the estimation elsewhere. 
\end{enumerate}

\begin{algorithm}[!t]
\begin{algorithmic}
    \STATE {\bfseries Input:} 
    $K$ training sequences: $X = \{x_i^{(k)}\}_{i=1, \dots, \mathbb{N}_k(\mathcal{X}),\ k=1, \dots, K}$, learning epoch $E$, learning rate $\gamma$, mini-batch size $M$. \;
    \STATE {\bfseries Initialization:} model parameters $\theta$, $e = 0$\; 
    \WHILE{$e<E$}
        \FOR{each sampled batch $\widehat{X}^{M}$ with size $M$}
        \STATE 1. Draw samples $z$ from noise distribution $\mathcal{N}(0, 1)$;
        \STATE 2. Feed $z$ into the generator $g$ to obtain sampled events $\widetilde{x}$;
        \STATE 3. Estimate conditional PDF (\ref{eq:kde}) and log-likelihood $\ell$ (\ref{eq:sequence-likelihood});
        \STATE 4. $\theta \leftarrow \theta + \gamma \partial \ell / \partial \theta$;
        \ENDFOR
        \STATE $e \leftarrow e + 1$;
    \ENDWHILE
    \STATE {\bfseries return} $\theta$ \;
\end{algorithmic}
\caption{Non-parametric learning for \texttt{CEG}}
\label{alg:non-para-learning}
\end{algorithm}

\begin{figure}[!t]
    \centering
    \begin{subfigure}[h]{.5\linewidth}
        {\includegraphics[width=\linewidth]{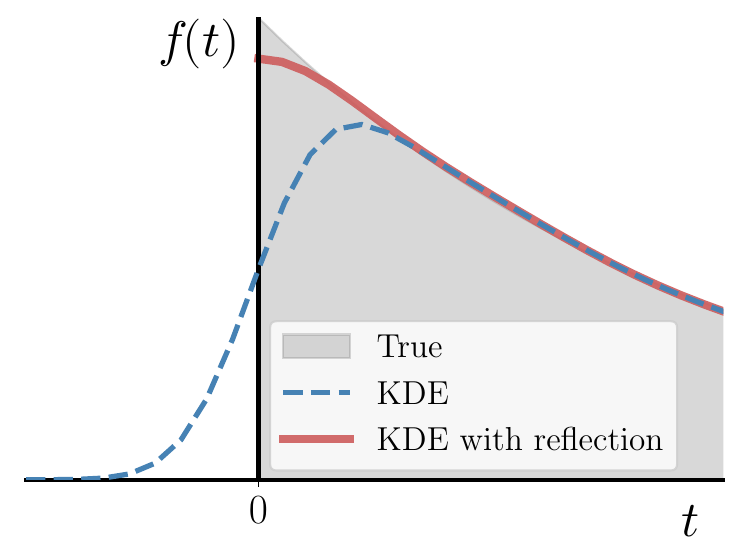}}
    \end{subfigure}
\caption{A comparison between the vanilla KDE and the KDE with boundary correction. The grey shaded area indicates the true density function, which is defined on the bounded region $[0, +\infty)$. The blue dashed line and red line show the estimated density function by the vanilla KDE and the KDE with reflection, respectively.}
\vspace{-0.05in}
\label{fig:illustrative-of-reflection-kde}
\end{figure}

\begin{figure}[!t]
    \centering
    \begin{subfigure}[h]{\linewidth}
        {\includegraphics[width=\linewidth]{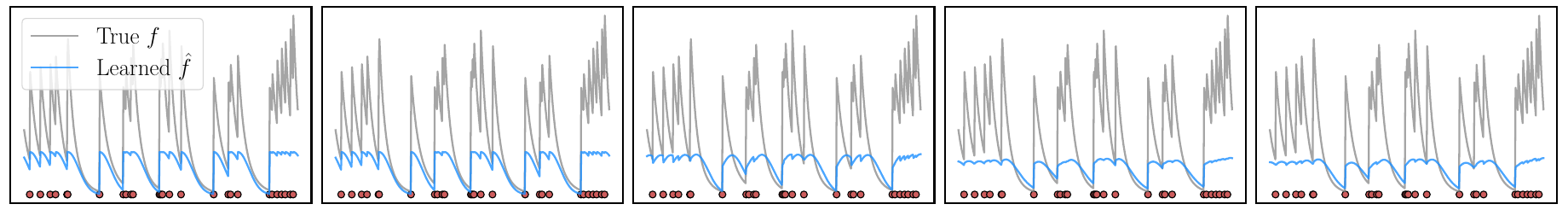}}
        \caption{Vanilla KDE}
        \end{subfigure}
    \vfill
    \begin{subfigure}[h]{\linewidth}
        {\includegraphics[width=\linewidth]{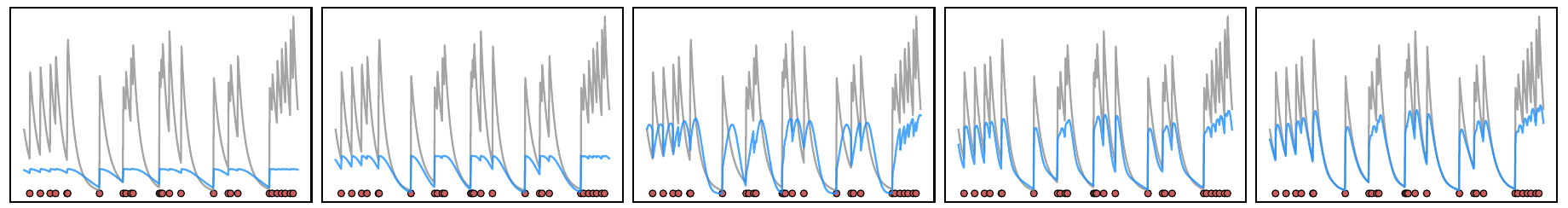}}
        \caption{KDE using self-tuned kernel}
    \end{subfigure}
    \vfill
    \begin{subfigure}[h]{\linewidth}
        {\includegraphics[width=\linewidth]{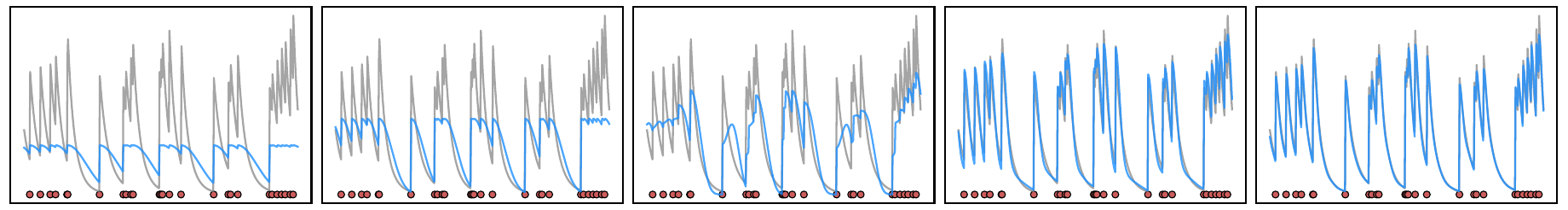}}
        \caption{KDE using self-tuned kernel with boundary correction}
    \end{subfigure}
\caption{The evolution of the estimated conditional PDF $f(t|\mathcal{H}_t)$ of an event sequence during training. Each row represents the model \texttt{CEG+KDE} trained with different variants of KDE. Each panel within the same row represents the estimated $f(t|\mathcal{H}_t)$ at every 10 training epochs.
}
\vspace{-0.05in}
\label{fig:example-of-kde}
\end{figure}

We present an illustrative example of how KDE with reflection benefits the boundary correction in Figure~\ref{fig:illustrative-of-reflection-kde}. 
It shows that the vanilla KDE extends the support of the density function to negative region and is biased near boundary (\ie, $t=0$), whereas KDE with reflection successfully corrects the boundary bias, leading to a more accurate estimation of the density function. 
Figure~\ref{fig:example-of-kde} compares the learned conditional PDF using three KDE methods on the same synthetic data set generated by a self-exciting Hawkes process. 
The results show that the estimation using the self-tuned kernel with boundary correction shown in (c) significantly outperforms two ablation models in (a) and (b). 
We also summarize the learning algorithm in Algorithm~\ref{alg:non-para-learning}.

\section{Variational learning}
\label{app:variational-learning}


The derivation of the ELBO in (\ref{eq:variational-lower-bound}) proceeds as follows. 
The conditional PDF can be re-written as:
$
    \log f_{\theta}(x|\boldsymbol{h}) = \log \int p_{\theta}(x, z|\boldsymbol{h})dz,
$
where $z$ is a latent random variable.
This integral has no closed form and can usually be estimated by Monte Carlo integration with importance sampling, \ie, 
$
    \int p_{\theta}(x, z|\boldsymbol{h})dz = \mathbb{E}_{z\sim q(\cdot |x, \boldsymbol{h})}\left[\frac{p_{\theta}(x, z|\boldsymbol{h})}{q(z|x, \boldsymbol{h})}\right].
$
Here $q(z|x, \boldsymbol{h})$ is the proposed variational distribution, where we can draw sample $z$ from this distribution given $x$ and $\boldsymbol{h}$. 
By Jensen's inequality, we can find the ELBO of the conditional PDF:
\[
    \begin{aligned}
        \log f_{\theta}(x|\boldsymbol{h}) 
        \geq \mathbb{E}_{z\sim q(\cdot|x, \boldsymbol{h})} \left[ \log \frac{p_{\theta}(x, z|\boldsymbol{h})}{q(z|x, \boldsymbol{h})}\right].
    \end{aligned}
\]
Using Bayes rule, the ELBO can be equivalently expressed as:
\begin{align*}
    &\mathbb{E}_{z\sim q(\cdot|x, \boldsymbol{h})} \left[ \log \frac{p_{\theta}(x, z|\boldsymbol{h})}{q(z|x, \boldsymbol{h})}\right] 
    =~ \mathbb{E}_{z\sim q(\cdot|x, \boldsymbol{h})} \left[ \log \frac{p_{\theta}(x|z, \boldsymbol{h})p_{\theta}(z|\boldsymbol{h})}{q(z|x, \boldsymbol{h})}\right] \\ 
    &= \mathbb{E}_{z\sim q(\cdot|x, \boldsymbol{h})} \left[ \log \frac{p_{\theta}(z|\boldsymbol{h})}{q(z|x, \boldsymbol{h})}\right] + \mathbb{E}_{z\sim q(\cdot|x, \boldsymbol{h})} \left[ \log p_{\theta}(x|z, \boldsymbol{h}) \right] \\
    &= -D_\text{KL}(q(z|x, \boldsymbol{h}) || p_{\theta}(z|\boldsymbol{h})) + \mathbb{E}_{z\sim q(\cdot|x, \boldsymbol{h})} \left[ \log p_{\theta}(x|z, \boldsymbol{h}) \right].
\end{align*}


\begin{algorithm}[!t]
\begin{algorithmic}
    \STATE {\bfseries Input:} 
    $K$ training sequences: $X = \{x_i^{(k)}\}_{i=1, \dots, \mathbb{N}_k(\mathcal{X}),\ k=1, \dots, K}$, learning epoch $E$, learning rate $\gamma$, mini-batch size $M$. \;
    \STATE {\bfseries Initialization:} model parameters $\theta$, $e = 0$\; 
    \WHILE{$e<E$}
        \FOR{each sampled batch $\widehat{X}^{M}$ with size $M$}
        \STATE 1. Draw samples $\epsilon$ from noise distribution $\mathcal{N}(0, 1)$;
        \STATE 2. Compute $z$ using reparametrization trick, given data $\widehat{X}^{M}$, noise $ \epsilon$, generator $g$, and $g_{\text{encode}}$;
        \STATE 3. Compute ELBO (\ref{eq:variational-lower-bound}) and log-likelihood $\ell$ (\ref{eq:sequence-likelihood});
        \STATE 4. $\theta \leftarrow \theta - \gamma \partial \ell / \partial \theta$;
        \ENDFOR
        \STATE $e \leftarrow e + 1$;
    \ENDWHILE
    \STATE {\bfseries return} $\theta$ \;
    
\end{algorithmic}
\caption{Variational learning for \texttt{CEG}}
\label{alg:variational-learning}
\end{algorithm}

In practice, we parameterize both $g_\text{encode}$ and the event generator using fully-connected neural networks with one hidden layer. The learning algorithm has been summarized in Algorithm~\ref{alg:variational-learning}.


\section{Learning with CDDM}
\label{app:conditional-ddpm}

The forward process in CDDM is fixed to a Markov chain to characterize the variational distribution of latent variables:
\[
    \begin{aligned}
        &q(z_{1:K}|x, \bm{h}) \coloneqq  \prod_{k=1}^{K} q(z_{k}|z_{k-1}, \bm{h}), \quad\\ &q(z_{k}|z_{k-1}, \bm{h}) \coloneqq \mathcal{N}(\sqrt{1 - \beta_{k}}z_{k-1}, \beta_{k}I).
    \end{aligned}
\]
Here, we define $z_0 \coloneqq x$. The forward process gradually adds noise to the data, and the final latent variable $z_K$ has the noise distribution of $\mathcal{N}(0, I)$. The variance schedule $\{\beta_k\}_{k=1}^{K}$ can be learned by reparameterization \citep{kingma2013auto} or held constant as hyperparameters \citep{ho2020denoising}. 
A reverse process inverts the transformation from the noise distribution with learned Gaussian transitions:
\[
    \begin{aligned}
        &p_{\theta}(x, z_{1:K}|\bm{h}) \coloneqq p(z_{K}) \prod_{k=1}^{K} p_{\theta}(z_{k-1}|z_{k},\bm{h}), \quad\\ &p_{\theta}(z_{k-1}|z_{k},\bm{h}) \coloneqq \mathcal{N}(\mu_{\theta}(z_{k}, k), \Sigma_{\theta}(z_{k}, k)),
    \end{aligned}
\]
where the mean $\mu_{\theta}$ is represented by a neural network and the covariance $\Sigma_{\theta}$ is set to be $\beta_k I$. 
The $\mu_{\theta}$ can be trained by maximizing the variational bound on the log-likelihood of the data $x$:
\[
    \mathbb{E}[\log f_{\theta}(x|\bm{h})] \geq \mathbb{E}_q\left[ \log p(z_K) + \sum_{k\geq1}\log\frac{p_{\theta}(z_{k-1}|z_k,\bm{h})}{q(z_k|z_{k-1},\bm{h})} \right].
\]
The property of the forward process allows the closed-form distribution of $z_k$ at an arbitrary step $k$ given the data $x$: denoting $\alpha_k \coloneqq 1-\beta_k$ and $\bar{\alpha}_k \coloneqq \prod_{r=1}^k\alpha_r$, we have $q(z_k|x,\bm{h}) = \mathcal{N}(\sqrt{\bar{\alpha}_k}x, (1-\bar{\alpha}_k)I)$.
The variational bound can be therefore expressed by the KL divergence between Gaussian distributions of $z_{1:K}$ and $x$, leading to the training objective of 
$\mathbb{E}_q\left[ \sum_{k>1}\frac{1}{2\beta_k}\|\tilde{\mu}_k(z_k, x) - \mu_{\theta}(z_k, k)\|^2 \right] + C$. Given the closed-form distribution of $z_k$ given $x$, we can re-parametrize $z_k = \sqrt{\bar{\alpha}_k}x + \sqrt{1-\bar{\alpha}_k}\epsilon$ with $\epsilon\sim \mathcal{N}(0, I)$. The model is trained on the final variant of the variational bound:
\[
    \mathbb{E}_{k, x, \epsilon}\left[ \| \epsilon - \epsilon_{\theta}(\sqrt{\bar{\alpha}_k}x + \sqrt{1-\bar{\alpha}_k}\epsilon, \bm{h}, k)\|^2 \right].
\]

\paragraph{Implementation details}
The core idea of classifier-free diffusion guidance is to simultaneously train an unconditional denoising diffusion model $\epsilon'_{\theta}(\sqrt{\bar{\alpha}_k}x + \sqrt{1-\bar{\alpha}_k}\epsilon, k)$, which later on is used to adjust the sampling steps together with the conditional denoising model. In practice, we can use a single neural network to parameterize both $\epsilon'_{\theta}$ and $\epsilon_{\theta}$, by simply input a null token $\varnothing$ in the place of the history embedding $\bm{h}$ for the unconditional model. Therefore, we have $\epsilon'_{\theta} = \epsilon_{\theta}(\sqrt{\bar{\alpha}_k}x + \sqrt{1-\bar{\alpha}_k}\epsilon, \varnothing, k)$. 
Algorithm~\ref{alg:CDDM-training} and Algorithm~\ref{alg:generator-g} show the training and sampling procedure, respectively.

\begin{algorithm}[!t]
\begin{algorithmic}
    \STATE {\bfseries Input:} 
    $K$ training sequences: $X = \{x_i^{(k)}\}_{i=1, \dots, \mathbb{N}_k(\mathcal{X}),\ k=1, \dots, K}$, learning epoch $E$, learning rate $\gamma$, mini-batch size $M$. \;
    \STATE {\bfseries Initialization:} model parameters $\theta$, $e = 0$\; 
    \WHILE{$e<E$}
        \FOR{each sampled batch $\widehat{X}^{M}$ with size $M$}
        \STATE 1. Draw samples $(x, \bm{h})$; set $\bm{h} \leftarrow \varnothing$ with probability $0.1$, which is pre-chosen for unconditional training;
        \STATE 3. Sample $k \sim \text{Unif}(1, K), \epsilon \sim \mathcal{N}(0, I)$;
        \STATE 4. Compute $\ell = \|\epsilon - \epsilon_{\theta}(\sqrt{\bar{\alpha}_k}x + \sqrt{1-\bar{\alpha}_k}\epsilon, \bm{h}, k)\|^2$
        \STATE 5. $\theta \leftarrow \theta - \gamma \partial \ell / \partial \theta$;
        \ENDFOR
        \STATE $e \leftarrow e + 1$;
    \ENDWHILE
    \STATE {\bfseries return} $\theta$ \;
    
\end{algorithmic}
\caption{Training of \texttt{CDDM} with classifier-free guidance}
\label{alg:CDDM-training}
\end{algorithm}

\section{Experiment details}
\label{append:additional-results}

\paragraph{Synthetic data description}
\label{app:synthetic-data}



\begin{algorithm}[!t]
\begin{algorithmic}
    \STATE {\bfseries Input:} Model $\lambda(\cdot)$, time horizon $T$, mark space $\mathcal{M}$, Intensity upper bound $\Bar{\lambda}$, $\mathcal{H}_T = \emptyset, t=0, i=0$\; 
    \WHILE{$t<T$}
        \STATE 1. Sample $u \sim \text{Unif}(0, 1), m \sim \text{Unif}(\mathcal{M}), D \sim \text{Unif}(0, 1)$. \;
        \STATE 2. Compute $t \leftarrow t - \ln u / \Bar{\lambda}, \ \lambda = \lambda(t, m|\mathcal{H}_T)$. \;
        \IF{$D\Bar{\lambda} \leq \lambda$}
            \STATE $i \leftarrow i + 1$; $t_i = t, m_i = m$, $\mathcal{H}_T \leftarrow \mathcal{H}_T \cup \{(t_i, m_i)\}$. \;
        \ENDIF
    \ENDWHILE
    \STATE {\bfseries return} $\mathcal{H}_T \cap T \times \mathcal{M}$ \;
    
\end{algorithmic}
\caption{Thinning algorithm}
\label{alg:thinning-algorithm}
\end{algorithm}

We use the following point process models to generate 1D synthetic data sets using Algorithm \ref{alg:thinning-algorithm}:
\begin{enumerate}
    \item Self-exciting: $\lambda(t) = \mu + \sum_{t_i \in \mathcal{H}_t}\beta e^{-\beta(t-t_i)}, \mu = 0.1, \beta = 0.1$.

    \item Self-correcting: $\lambda(t) = \exp{\left(\mu t - \sum_{t_i \in \mathcal{H}_t}\alpha\right)}, \mu = 1.0, \alpha = 1.0$.

    \item \textsf{T-MNIST}: In the MNIST series, all the digits that are greater than nine will be truncated to nine. The kernel for the event times are $k(t, t_i) = \beta e^{-\beta(t-t_i)}, \beta = 0.2$.
    
    \item \textsf{T-CIFAR} includes seven image types: bicycles/motorcycles (exercises), apples/pears/oranges (food ingestion), keyboards (work), and beds (sleeping).
    Before 21:00, the activity series progresses with the transition probability matrix between exercise, food ingestion, and work being 
    \[
    P = 
        \begin{pmatrix}
        0.0 & 1.0 & 0.0 \\
        0.2 & 0.0 & 0.8 \\
        0.2 & 0.3 & 0.5 \\
        \end{pmatrix}.
    \]
    After 21:00, the probability of sleeping increases linearly from 0 to 1 at 23:00. Each series ends with sleeping. The self-correcting process for event times is set with $\mu = 0.1, \alpha = 0.5$, indicating that each activity will last for a while before the next activity. 
\end{enumerate}

\paragraph{Experimental setup} 
Our score function $\epsilon_{\theta}$ is a fully-connected neural network with three hidden layers of width $128$ with softplus activation function. We transform the generated times with a softplus function $\text{Softplus}(x) = \frac{1}{\beta}\log \left(1 + e^{\beta x}\right)$ to ensure the time interval is always positive.
For \texttt{RMTPP}, \texttt{NH} and \texttt{FullyNN}, we take the default hyper-parameters in the original papers, with the dimension of history embedding being $64$. We use a fully-connected neural network with two hidden layers of width $64$ for the cumulative hazard function in \texttt{FullyNN}. No hyperparameter in \texttt{ETAS}.
The training/testing ratio is $9:1$ for all models.
We maximize log-likelihood (\ref{eq:sequence-likelihood}) using Adam optimizer \citep{kingma2014adam} with a learning rate of $10^{-3}$ and a batch size of 32 (event sequences). 
Experiments are implemented on Google Colaboratory (Pro) with 12GB RAM and a Tesla T4 GPU.

The predictive and discriminative score \citep{yoon2019time} are adopted to evaluate the fidelity and diversity of the generated time series samples:
\begin{itemize}
    \item Predictive score: we train a sequence-prediction model (usually a 2-layer LSTM) using the generated data and evaluate it on the original data in terms of the mean absolute error of event time prediction. A lower score indicates a better reproduction of the temporal properties of the original data by the generated data. 

    \item Discriminative score: we train a time-series classification model (usually a 2-layer LSTM) to distinguish real and generated samples. We use 80\% of all data as the training set and evaluate it on the rest 20\%. A lower score (classification accuracy) means a higher similarity between real and generated samples.
\end{itemize}

\end{document}